\documentclass[11pt,a4paper]{article}

\usepackage[hyperref]{emnlp2018}

\usepackage{times}
\usepackage{epsfig}
\usepackage{graphicx}
\usepackage{amsmath}
\usepackage{amssymb}
\usepackage{breqn}
\usepackage[bottom]{footmisc}

\usepackage{latexsym}

\usepackage{url}

\usepackage[]{algorithm}
\usepackage[]{algpseudocode}

\makeatletter
\def\BState{\State\hskip-\ALG@thistlm}
\makeatother

\newcommand{\eadd}{\mathbin{\text{$\vcenter{\hbox{\textcircled{$+$}}}$}}}

\aclfinalcopy % Uncomment this line for the final submission

\title{Multimodal Differential Network for Visual Question Generation}

  \author{ $\textbf{Badri N. Patro}$  \quad $\textbf{Sandeep Kumar}$ \quad $\textbf{Vinod K. Kurmi} $ \quad  $\textbf{Vinay P. Namboodiri}$ \\
  Indian Institute of Technology, Kanpur \\
  {\tt \{badri,sandepkr,vinodkk,vinaypn\}@iitk.ac.in} \\
  \\}

\date{}

\begin{document}
\maketitle

\begin{abstract}

Generating natural questions from an image is a semantic task that requires using visual and language modality to learn multimodal representations. Images can have multiple visual and language contexts that are relevant for generating questions namely places, captions, and tags. In this paper, we propose the use of exemplars for obtaining the relevant context.
%Rather than relying on explicit context tags obtained through exemplars, we consider implicit context obtained by learning an embedding. 
We obtain this by using a Multimodal Differential Network to produce natural and engaging questions. The generated questions show a remarkable similarity to the natural questions as validated by a human study. 
% Ablation studies show that the proposed architecture is most suited for the question generation task.
Further, we observe that the proposed approach substantially improves over state-of-the-art benchmarks on the quantitative metrics (BLEU, METEOR, ROUGE, and CIDEr).

\end{abstract}
%~\citep{Aho:72}

\section{Introduction}\label{intro}
%%%%%%%%%%%%%%%%%%%%%%%%%%%%%%%%%%%%%%%%%%%%%%%%%%%%%%%%%%%%%%%%%%%%%%%
%Interaction of humans and automated systems is an important and increasingly active area of research. One such aspect is based on vision and language based interaction. This area has seen a number of works related to visual question answering~\cite{VQA} and visual dialog~\cite{visdial}. Based on work aimed at autonomous interaction of robots for visual dialog, it has emerged that it is extremely important for autonomous agents to learn to ask natural questions about an image ~\cite{visdial_rl}. While not as well studied as the other tasks of answering questions or carrying a conversation, there has been work aimed at generating natural questions from an image~\cite{Mostafazadeh_ACL2016, jain_CVPR2017}. The underlying principle for all these methods is an encoder-decoder formulation. We argue that there are underlying cues that motivate a natural question about an image. It is important to incorporate the best cue among these while generating questions.
To understand the progress towards multimedia vision and language understanding, 
%  that is possible through interesting tasks such as image captioning,
 a visual Turing test was proposed by~\cite{Geman_PNAS2015} that was aimed at visual question answering~\cite{VQA}. Visual Dialog~\cite{visdial} is a natural extension for VQA. Current dialog systems as evaluated in~\cite{visdial_eval} show that when trained between bots, AI-AI dialog systems show improvement, but that does not translate to actual improvement for Human-AI dialog. This is because, the questions generated by bots are not natural (human-like) and therefore does not translate to improved human dialog. Therefore it is imperative that improvement in the quality of questions will enable dialog agents to perform well in human interactions. Further, ~\cite{GanjuCVPR17} show that unanswered questions can be used for improving VQA, Image captioning and Object Classification. 
% So generation of natural questions will further improve performance on these tasks.

\begin{figure}[ht]
	%\vspace{1in}
	\centering
	\includegraphics[width=0.5\textwidth]{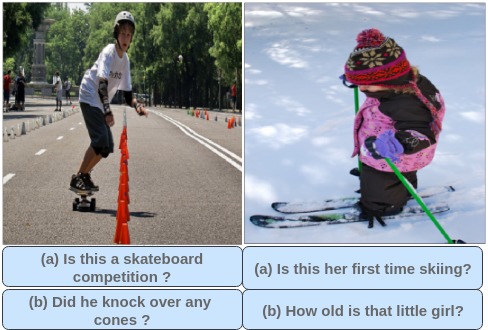}
	\vspace{-0.7cm}
	\caption{Can you guess which among the given questions is human annotated and which is machine generated? \footnotemark[0] }
	\label{fig:vqg_result}
\end{figure}

%  This task has seen considerable progress since then.
%  with many datasets aimed at solving this problem.
%  However, to obtain real interaction between humans and automated systems, it is important that the system also learns to ask questions~\cite{visdial_rl}.
 \noindent An interesting line of work in this respect is the work of~\cite{mostafazadeh2016generating}. Here the authors have proposed the challenging task of generating {\em natural} questions for an image.  %While generating answers provided an image and a question could be solved through correlation of features, the task of asking natural questions is much harder~\cite{Mostafazadeh_ACL2016}. %  To solve this, there have been a few works that aim to learn sequential encoder-decoder models~\cite{Mostafazadeh_ACL2016} or variational encoder-decoder models~\cite{jain_CVPR2017}.
% One aspect that is central to a question is the context that is relevant to generate it. However, this context changes for every image. An image with a train or bus would lead to questions about journey whereas images that depict some game being played could have other questions such as, ``Which team is winning?''. How can one have widely varying context provided for generating questions? To solve this problem, we use the context obtained by considering exemplars, specifically we use the difference between relevant and irrelevant exemplars. We consider different contexts in the form of Location, Caption, and Part of Speech tags and caption outperformed others as shown in the Ablation analysis.
One aspect that is central to a question is the context that is relevant to generate it. However, this context changes for every image. As can be seen in Figure~\ref{fig:vqg_result}, an image with a person on a skateboard would result in questions related to the event. Whereas for a little girl, the questions could be related to age rather than the action. %An image with a train or bus would lead to questions about journey whereas images that depict some game being played could have other questions such as, ``Which team is winning?''. 
How can one have widely varying context provided for generating questions? To solve this problem, we use the context obtained by considering exemplars, specifically we use the difference between relevant and irrelevant exemplars. We consider different contexts in the form of Location, Caption, and Part of Speech tags.
% An aspect that is not included in these previous attempts is to consider that in asking a natural question, fine-grained information that captures the difference between embeddings is required to ensure that the right context is captured. 
%In this paper we propose a multimodal differential network that provides us with embeddings that helps us in tackling this problem. 
% \begin{figure}[ht]
% 	%\vspace{1in}
% 	\centering
% 	\includegraphics[width=0.5\textwidth]{fig/MDN_VQG.pdf}
% 	\vspace{-0.7cm}
% 	\caption{Can you guess which among the given questions is human annotated and which is machine generated? The human annotated questions are (b) for the first image and (a) for the second image.}
% 	\label{fig:vqg_result}
% \end{figure}
\footnotetext{The human annotated questions are (b) for the first image and (a) for the second image.}

\begin{figure}[ht]
%\vspace{1in}
\centering
\includegraphics[width=0.5\textwidth]{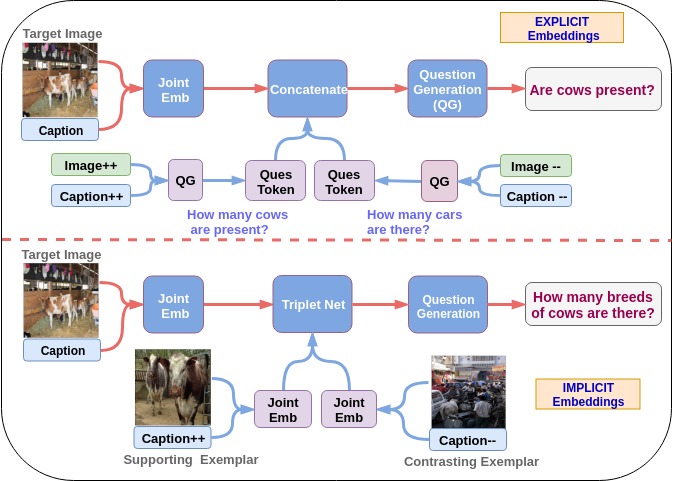}
\vspace{-0.7cm}
\caption{Here we provide intuition for using implicit embeddings instead of explicit ones. As explained in section~\ref{intro}, the question obtained by the implicit embeddings are natural and holistic than the explicit ones.}
\label{fig:exemplar_v1}
\end{figure}

\noindent Our method implicitly uses a differential context obtained through supporting  and contrasting exemplars to obtain a differentiable embedding. This embedding is used by a question decoder to decode the appropriate question. As discussed further, we observe this implicit differential context to perform better than an explicit keyword based context. The difference between the two approaches is illustrated in  Figure~\ref{fig:exemplar_v1}. This also allows for better optimization as we can backpropagate through the whole network. We provide detailed empirical evidence to support our hypothesis. As seen in Figure~\ref{fig:vqg_result} our method generates natural questions and improves over the state-of-the-art techniques for this problem. 
% An example of the natural question generation is illustrated in figure~\ref{fig:vqg_result}.
% As can be seen, the method can generate extremely natural questions.

% \vspace{-0.7cm}
%A challenge we face is in obtaining semantic exemplars (typical examples). The differential network relies on learning the difference in multimodal context between near and far samples. Therefore, the exemplars should be meaningful. We obtain exemplars by considering the joint image-caption based embedding. The differential context in the joint image-caption embedding space is learned. We evaluate different methods for obtaining the joint multimodal space for the embeddings such as concatenation, Hadamard product etc.  We observe that the Joint method (concatenation) that learns a joint map provides improved performance for this task. These multimodal embeddings are trained to differentiate between near and far examples using a triplet learning method. These embeddings are then used by a decoder and are trained to generate natural questions. Through this paper we provide the following contributions

To summarize, we propose a multimodal differential network to solve the task of visual question generation. Our contributions are: (1) A method to incorporate exemplars to learn differential embeddings that captures the subtle differences between supporting and contrasting examples and aid in generating natural questions. (2) We provide Multimodal differential embeddings, as image or text alone does not capture the whole context and we show that these embeddings outperform the ablations which incorporate cues such as only image, or tags or place information. (3) We provide a thorough comparison of the proposed network against state-of-the-art benchmarks along with a user study and statistical significance test.
%  \begin{itemize}
%  \item A method to incorporate exemplars to learn differential embeddings that captures the subtle differences between near and far examples and aid in generating natural questions.
%  \item We provide Multimodal differential embeddings, as image or text alone does not capture the whole context and we show that these embeddings outperform the ablations which incorporate cues such as only image, or tags or place information.
%  \item  We also provide a thorough comparison of the proposed network against state-of-the-art benchmarks on standard datasets along with a detailed analysis in terms of a user study and statistical significance test.
%   \end{itemize} 

%%%%%%%%%%%%%%%%%%%%%%%%%%%%%%%%%%%%%%%%%%%%%%%%%%%%%%%%%%%%%%%%%%%%%%%
\section{Related Work}
%%%%%%%%%%%%%%%%%%%%%%%%%%%%%%%%%%%%%%%%%%%%%%%%%%%%%%%%%%%%%%%%%%%%%%%

% \begin{figure*}[ht]
% %\vspace{1in}
% \centering
% %\includegraphics[width=0.7\textwidth]{fig/VQA_MDN_new.png}
% \includegraphics[width=1.0\textwidth]{fig/exemplar_v1.pdf}
% \vspace{-0.7cm}
% \caption{In this figure we provide intuition for using implicit embeddings instead of explicit ones. As explained in section~\ref{intro}, the question obtained by the implicit embeddings are more natural and holistic than the explicit ones.}
% \label{fig:exemplar_v1}
% \end{figure*}

\label{sec:lit_surv}
Generating a natural and engaging question is an interesting and challenging task for a smart robot (like chat-bot). It is a step towards having a natural visual dialog instead of the widely prevalent visual question answering bots. Further, having the ability to ask natural questions based on different contexts is also useful for artificial agents that can interact with visually impaired people. While the task of generating question automatically is well studied in NLP community, it has been relatively less studied for image-related natural questions. This is still a difficult task~\cite{mostafazadeh2016generating} that has gained recent interest in the community.
    
    % In the language community,~\cite{Rus_ACL2010} introduced the shared evaluation for the task of generating questions from paragraph and sentences.~\cite{Heilman_ACL2010} proposed an over-generated question based on general purpose syntactic mapping. They generated simple declarative sentences from a given paragraph and then converted it into a candidate question by using syntactic transformations. There have been many rule-based approaches for text-based question generation such as~\cite{Mazidi_ACL2014}. The problem with rule-based approaches are that they are more syntactic rather than semantic.~\cite{Chali_CL2015} developed a model for generating all possible questions using topic modeling and named entity relation.
Recently there have been many deep learning based approaches as well for solving the text-based question generation task such as~\cite{Du_ArXiv2017}.
Further,~\cite{Serban_Arxiv2016} have proposed a method to generate a factoid based question based on triplet set \{subject, relation and object\} to capture the structural representation of text and the corresponding generated question.

These methods, however, were limited to text-based question generation. There has been extensive work done in the Vision and Language domain for solving image captioning, paragraph generation, Visual Question Answering (VQA) and Visual Dialog.~\cite{Barnard_JMLR2003,Farhadi_ECCV2010,Kulkarni_CVPR2011} proposed conventional machine learning methods for image description.~\cite{Socher_TACL2014,Vinyals_CVPR2015,Karpathy_CVPR2015,Xu_ICML2015,Fang_CVPR2015,Chen_CVPR2015,Johnson_CVPR2016,Yan_ECCV2016} have generated descriptive sentences from images with the help of Deep Networks.
There have been many works for solving Visual Dialog~\cite{chappell_HFES2004,Das_EMNLP2016,visdial,de2017guesswhat,strub2017end}. 
A variety of methods have been proposed by~\cite{Malinowski_NIPS2014,Lin_ECCV2014,VQA,Ren_NIPS2015,Ma_AAAI2016,Noh_CVPR2016} for solving VQA task including attention-based methods~\cite{Zhu_CVPR2016,Fukui_arXiv2016,Gao_NIPS2015,Xu_ECCV2016,Lu_NIPS2016,Shih_CVPR2016,Patro_CVPR2018}.
% There has been significant interest in including attention for the VQA problem which was introduced by~\cite{Zhu_CVPR2016,Fukui_arXiv2016,Gao_NIPS2015,Xu_ECCV2016,Lu_NIPS2016,Shih_CVPR2016}.
% These were obtained by manually generated questions and could have benefited by a Visual Question Generation module due to the ability of creating a large number of questions for any image and would also have reduced human effort. 
However, Visual Question Generation (VQG) is a separate task which is of interest in its own right and has not been so well explored~\cite{mostafazadeh2016generating}. This is a vision based novel task aimed at generating natural and engaging question for an image.% using deep CNN. 
    % This was first proposed by~\cite{Mostafazadeh_ACL2016}.
   ~\cite{Yang_arXiv2015} proposed a method for continuously generating questions from an image and subsequently answering those questions.
    % In this paper we focus only on the question generation task that has been explored subsequently. 
    The works closely related to ours are that of~\cite{mostafazadeh2016generating} and~\cite{jain2017creativity}. In the former work, the authors used an encoder-decoder based framework whereas in the latter work,
    % In their work they have generated open ambiguous question for visual input, which has no determined answer for that question. 
    % In~\cite{jain_CVPR2017},
    the authors extend it by using a variational autoencoder based sequential routine to obtain natural questions by performing sampling of the latent variable.

%%%%%%%%%%%%%%%%%%%%%%%%%%%%%%%%%%%%%%%%%%%%%%%%%%%%%%%%%%%%%%%%%%%%%%%
%\subsection{Background}
%%%%%%%%%%%%%%%%%%%%%%%%%%%%%%%%%%%%%%%%%%%%%%%%%%%%%%%%%%%%%%%%%%%%%%%
\section{Approach}
\begin{figure}[ht]
%\vspace{1in}
\centering
\includegraphics[width=1\columnwidth]{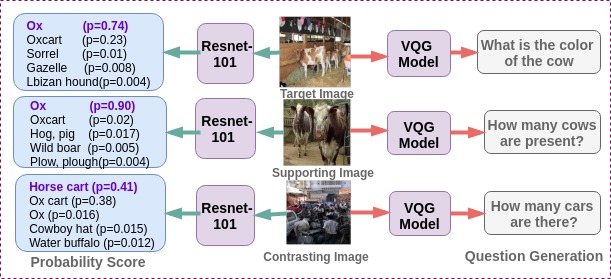}
\vspace{-0.7cm}
\caption{An illustrative example shows the validity of our obtained exemplars with the help of an object classification network, RESNET-101. We see that the probability scores of target and supporting exemplar image are similar. That is not the case with the contrasting exemplar. The corresponding generated questions when considering the individual images 
% generated (This VQG model only takes a single image as input) 
are also shown.  %a single image and its caption. We observe that the questions generated in figure~\ref{fig:exemplar_v1} are more natural as they are taking into consideration all these images.
}
\label{fig:exemplar}
\end{figure} 
%\subsubsection{Explicit vs Implicit Module}
% In this experiment, we select a target image and obtained corresponding supporting and contrasting exemplar and conducted a recognition task and question generation task. In recognition task, We have predicted the objects in the target image along with supporting and contrasting exemplar using pre trained image recognition model resnet-101\cite{He_CVPR2016} as shown in figure ~\ref{fig:exemplar}, which provider top five class with confidence score. We have observed that target and supporting image belong to similar class  where as the contrasting image are different class with larger margin of probability. 
\noindent In this section, we clarify the basis for our approach of using exemplars for question generation. To use exemplars for our method, we need to ensure that our exemplars can provide context and that our method generates valid exemplars. 

 We first analyze whether the exemplars are valid or not. We illustrate this in figure~\ref{fig:exemplar}. We used a pre-trained RESNET-101~\cite{He_CVPR2016} object classification network on the target, supporting and contrasting images. We observed that the supporting image and target image have quite similar probability scores. The contrasting exemplar image, on the other hand, has completely different probability scores.

 Exemplars aim to provide appropriate context. To better understand the context, we experimented by analysing the questions generated through an exemplar. We observed that indeed a supporting exemplar could identify relevant tags (cows in Figure~\ref{fig:exemplar}) for generating questions.
% We observed that indeed a supporting exemplar could capture context by identifying the tags (cows in Figure~\ref{fig:exemplar}) that are relevant for generating questions.
\noindent We improve use of exemplars by using a triplet network. This network ensures that the joint image-caption embedding for the supporting exemplar are closer to that of the
target image-caption and vice-versa.
We empirically evaluated whether an explicit approach that uses the differential set of tags as a one-hot encoding improves the question generation, or the implicit embedding obtained based on the triplet network.
We observed that the implicit multimodal differential network empirically provided better context for generating questions. Our understanding of this phenomenon is that both target and supporting exemplars generate similar questions whereas contrasting exemplars generate very different questions from the target question. The triplet network that enhances the joint embedding thus aids to improve the generation of target question. These are observed to be better than the explicitly obtained context tags as can be seen in Figure~\ref{fig:exemplar_v1}. 
% This is because the explicit tags do not capture the full context available and the differential embedding can be used by the question generator to obtain more meaningful questions.%We initially give the questions generated by only considering the target image and its caption. Similarly we provide the questions for only supporting and contrasting exemplars. Then we provide the question generated by our model and improvement in the quality of the question is seen. By providing these supporting and contrasting exemplars we are conditioning the model to generate a particular type of question.% In question generation, We have used our reference visual question generation model to generate the target question, supporting and contrasting  question. % will used to remove the confusion term in the image.%Basically . We also experimented with explicitly providing one-hot encoding of the question tokens obtained from a simple image-caption network for VQG (MDN without supporting and contrasting networks and triplet loss). This network used joint image-caption embedding of supporting and contrasting exemplars as its input. We compare this with implicitly providing this conditioning with the help of  supporting and contrasting nets in our MDN in figure~\ref{fig:vqg_result} and we see that the results are better in the latter case.% for the classes. 
% We thus ensured that in our approach the exemplars could generate meaningful questions and that the exemplars are valid. 
We now explain our method in detail.

% This also shows that the supporting exemplars are near to the original image and vice-versa.
%%%%%%%%%%%%%%%%%%%%%%%%%%%%%%%%%%%%%%%%%%%%%%%%%%%%%%%%%%%%%%%%%%%%%%%
\section{Method}
%%%%%%%%%%%%%%%%%%%%%%%%%%%%%%%%%%%%%%%%%%%%%%%%%%%%%%%%%%%%%%%%%%%%%%%
\label{sec:method}
\begin{figure*}[ht]
	%\vspace{1in}
	\centering
	\includegraphics[width=0.7\textwidth]{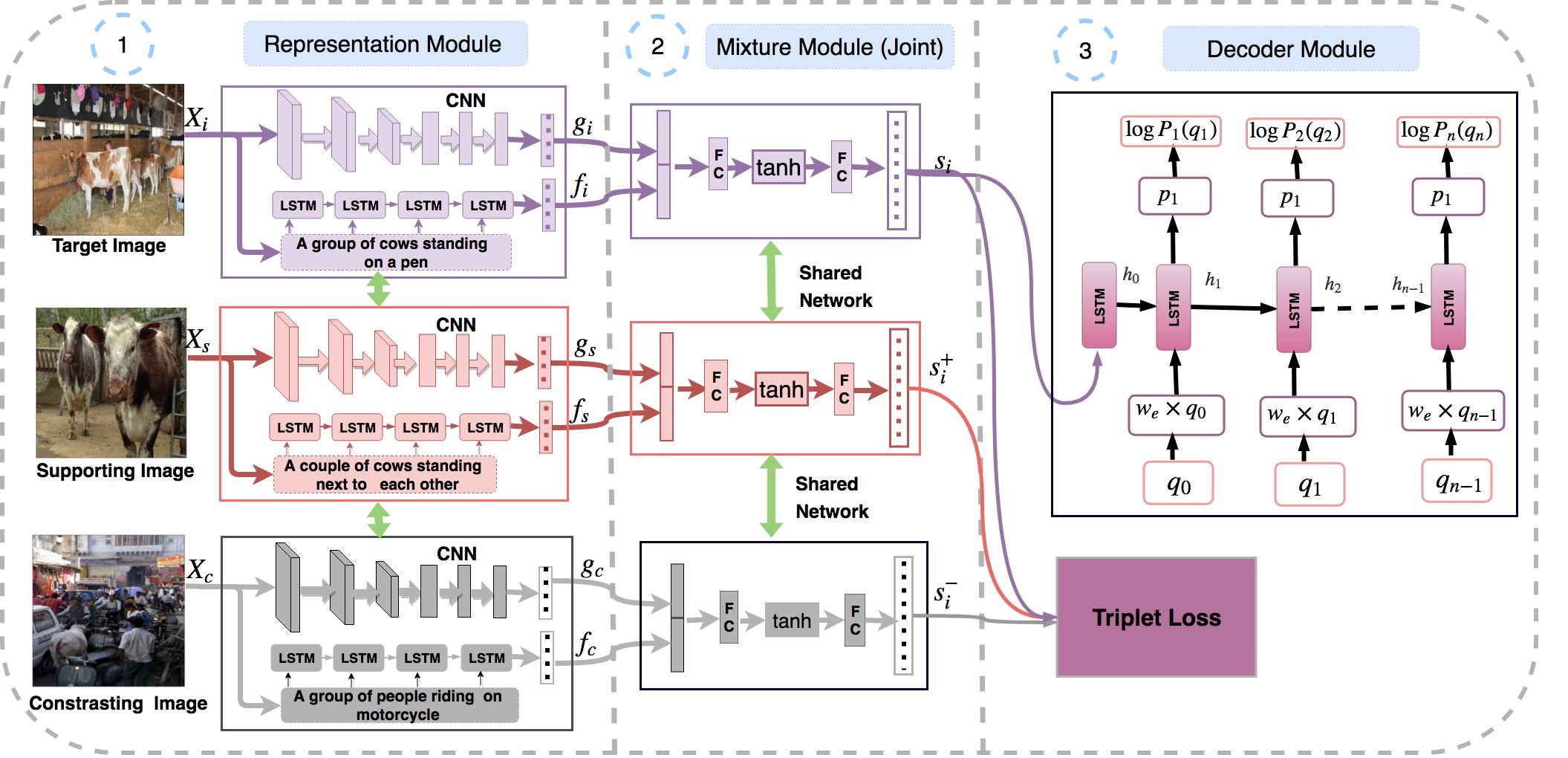}
	\vspace{-0.2cm}
	\caption{This is an overview of our Multimodal Differential Network for Visual Question Generation. It consists of a Representation Module which extracts multimodal features, a Mixture Module that fuses the multimodal representation and a Decoder that generates question using an LSTM based language model. In this figure, we have shown the Joint Mixture Module. We train our network with a Cross-Entropy and Triplet Loss.}
	\label{fig:MDN}
\end{figure*}

The task in visual question generation (VQG) is to generate a natural language question $\hat{Q}$, for an image $I$. We consider a set of pre-generated context $C$ from image $I$. We maximize the conditional probability of generated question given image and context as follows: % ques token
% \[  \hat{Q}=argmax_{Q \in \Omega}  log(P(Q |I ,C:\theta)) \] for inference equation
\begin{equation}
\label{theta_optim}
 \hat{\theta}=\arg\max_{\theta}\ \sum_{(I,C,Q)}{ \mbox{log}\ P( Q |\ I ,C,\theta )} 
\end{equation}
% \arg\max&{}\ 
where $\theta$ is a vector for all possible parameters of our model. $Q$ is the ground truth question. The log probability for the question is calculated by using joint probability over $\{{q_0},{q_1},.....,{q_N}\}$ with the help of chain rule. For a particular question, the above term is obtained as:
\begin{equation*}
 \mbox{log}\ P( \hat{Q} | I , C)=\sum_{t=0}^{N}{ \mbox{log}\ P( {q_t} | I ,C,{q_0},..,{q}_{t-1}} ) 
\end{equation*}
where $N$ is length of the sequence, and $q_t$ is the $t^{th}$ word of the question. We have removed $\theta$ for simplicity. 

Our method is based on a sequence to sequence network~\cite{Sutskever_NIPS2014,Vinyals_CVPR2015,Bahdanau_arXiv2014}. The sequence to sequence network has a text sequence as input and output. In our method, we take an image as input and generate a natural question as output. The architecture for our model is shown in Figure~\ref{fig:MDN}. Our model contains three main modules, (a) Representation Module that extracts multimodal features (b) Mixture Module that fuses the multimodal representation and (c) Decoder that generates question using an LSTM-based language model.

%  While training we learn the parameters for the model that optimizes equation~\ref{theta_optim}  over the whole training set. % considering each batch at a time using RMSPROP optimizer. The training data consist of thousands of images, captions of images and natural questions associated with each image. These human annotated questions are present in the  training dataset. 
%  The details regarding the training configuration and training dataset are provided in the supplementary material. At the time of inference, we sample a question word $q_i$ from the distribution with probability $p_i$ and continue sampling until the end token or maximum length for the question is reached. 
During inference, we sample a question word $q_i$ from the softmax distribution and continue sampling until the end token or maximum length for the question is reached. 
 We experimented with both sampling and \textit{argmax} and found out that \textit{argmax} works better. This result is provided in the supplementary material.
%%%%%%%%%%%%%%%%%%%%%%%%%%%%%%%%%%%%%%%%%%%%%%%%%%%%%%%%%%%%%%%%%%%%%%%
%%%%%%%%%%%%%%%%%%%%%%%%%%%%%%%%%%%%%%%%%%%%%%%%%%%%%%%%%%%%%%%%%%%%%%%
\subsection{Multimodal Differential Network}
%%%%%%%%%%%%%%%%%%%%%%%%%%%%%%%%%%%%%%%%%%%%%%%%%%%%%%%%%%%%%%%%%%%%%%%
The proposed Multimodal Differential Network (MDN) consists of a representation module and a joint mixture module. %For the Representation module we need the exemplars and their corresponding captions.

\subsubsection{Finding Exemplars}
% We use nearest neighbors for finding exemplars. A CNN is used to obtain image features as mentioned previously. We use a K-D Tree data structure to represent the features, and Euclidean metric is used for the distance between the features. 
% % For obtaining contrasting exemplars, a neighbor that is an order of magnitude further than the nearest neighbor was used.
% This is obtained through a coarse quantization of nearest neighbors of the training example into 50 clusters, and selecting the nearest as supporting and farthest as the contrasting exemplars.
We used an efficient KNN-based approach (k-d tree) with Euclidean metric to obtain the exemplars. This is obtained through a coarse quantization of nearest neighbors of the training examples into 50 clusters, and selecting the nearest as supporting and farthest as the contrasting exemplars. 
 We experimented with ITML based metric learning~\cite{davis_ACM2007} for image features. Surprisingly, the KNN-based approach outperforms the latter one. We also tried random exemplars and different number of exemplars and found that $k=5$ works best. We provide these results in the supplementary material.
% using the representative an exemplar as contrasting that was around 20 clusters away in a 50 cluster order.  This parameter is not stringent, and it only matters that the contrasting exemplar is relatively far from the supporting exemplar.

\subsubsection{Representation Module}
We use a triplet network~\cite{Frome_ICCV2007,Hoffer_Springer2015} in our representation module. 
% In literature , \cite{Patro_CVPR2018} is used a similar kind of network for VQA task.
We refereed a similar kind of work done in \cite{Patro_CVPR2018} for building our triplet network. %to obtain the joint context encoding vector based on the image and its exemplars. 
The triplet network consists of three sub-parts: target, supporting, and contrasting networks. All three networks share the same parameters. Given an image $x_i$ we obtain an embedding $g_i$ using a CNN parameterized by a function $G(x_i,W_c)$ where $W_c$ are the weights for the CNN. The caption $C_i$ results in a caption embedding $f_i$ through an LSTM parameterized by a function $F(C_i, W_l)$ where $W_l$ are the weights for the LSTM. This is shown in part 1 of Figure~\ref{fig:MDN}. %The output image embedding $g_i$ and caption embedding $f_i$ are used in a joint Mixture module(section ~\ref{mixture_model}) that combines the image and caption embedding with a weighted function and produces an output context vector $s_i$. The Joint mechanism is illustrated in figure~\ref{fig:MDN}. 
Similarly we obtain image embeddings $g_s$ \& $g_c$  and  caption embeddings $f_s$ \& $f_c$.  
\begin{equation}
 \begin{split}
&g_i=G(x_i,W_c)=CNN(x_i)\\
&f_i=F(C_i, W_l)=LSTM(C_i)
\end{split}
\end{equation}

%%%%%%%%%%%%%%%%%%%%%%%%%%%%%%%%%%%%%%%%%%%%%%%%%%%%%%%%%%%%%%%%%%%%%%%%%%%%
\subsubsection{Mixture Module}\label{mixture_model}
%%%%%%%%%%%%%%%%%%%%%%%%%%%%%%%%%%%%%%%%%%%%%%%%%%%%%%%%%%%%%%%%%%%%%%%
The Mixture module brings the image and caption embeddings to a joint feature embedding space. The input to the module is the embeddings obtained from the representation module. We have evaluated four different approaches for fusion viz., joint, element-wise addition, hadamard and attention method. Each of these variants receives image features $g_i$ \& the caption embedding $f_i$, and outputs a fixed dimensional feature vector $s_i$.  
The Joint method concatenates $g_{i}$  \& $f_{i}$ and maps them to a fixed length feature vector $s_{i}$ as follows:
\begin{equation}
     s_{i} =  W^{T}_{j} *  \tanh(  W_{ij}  g_{i} \ ^\frown \ (W_{cj}  f_{i} + b_j))
\end{equation}
% * is multiplication
where $g_{i}$ is  the 4096-dimensional  convolutional feature from the  FC7 layer of pretrained VGG-19 Net~\cite{simonyan_arXiv2014}. $W_{ij}, W_{cj},W_{j}$ are the weights and $b_j$ is the bias for different layers. $^\frown$ is the concatenation operator. 

% {Hadamard method} uses element-wise multiplication whereas {Addition method} uses element-wise addition in place of the concatenation operator of the Joint method. The Hadamard method finds a correlation between image feature and caption feature vector while the Addition method learns a resultant vector.
% In the attention method, the output $S_{i}$ is the weighted average of attention probability vector $P_{att}$ and convolutional features $G_{img}$. The attention probability vector weights the contribution of each convolutional feature based on the caption vector. This attention method is similar to work stack attention method~\cite{Yang_CVPR2016}. The attention mechanism is given by:
% \begin{equation}
%     \begin{split}
%         & h_{att}= \tanh({W_I}{G_{img}} \oplus ({W_C}{F_{cap}}+{b_c})) \\
%         & P_{att}= \mbox{Softmax}({W^T_P}{h_{att}}+{b_P}) \\
%         & V_{att}= \sum_{i}{P_{att}(i)}{G_{img}(i)}\\
%         & A_{att}= V_{att} + f_{i} \\
%         & s_{i}=\tanh({W_A} A_{att} + b_A)
%     \end{split}
% \end{equation}
% where $G_{img}$ is  the 14x14x512-dimensional  convolution feature map from the fifth convolution layer of VGG-19 Net of image $X_{i}$ and $f_{i}$ is the caption context vector. The attention probability vector $P_{att}$ is a 196-dimensional vector. $W_{I},W_{C},W_{P}$ are the weights and $b_c, b_A, b_c $ is the bias for different layers. We evaluate the different approaches and provide results for the same. Here $\oplus$ represents element-wise addition.

 Similarly, We obtain context vectors  $s^+_i$ \& $s^-_i$ for the supporting and contrasting exemplars. Details for other fusion methods are present in supplementary.% The aim of the triplet network~\cite{Schroff_CVPR2015} is to obtain context vectors that bring the supporting exemplar close to the target image and far from the contrasting exemplar.
The aim of the triplet network~\cite{Schroff_CVPR2015} is to obtain context vectors that bring the supporting exemplar embeddings closer to the target embedding and vice-versa.
% pushes the contrasting exemplar embeddings far away from the target embeddings. 
This is obtained as follows:
\begin{equation}
 \begin{split}
& D(t(s_{i}),t(s_{i}^{+})) +\alpha < D(t(s_{i}),t(s_{i}^{-}))\\
& \forall{(t(s_{i}),t(s_{i}^{+}),t(s_{i}^{-}))} \in M,
\end{split}
\end{equation}
where $D(t(s_{i}),t(s_{j})) = ||t(s_{i})- t(s_{j})||_{2}^{2}$ is the euclidean distance between two embeddings $t(s_{i})$ and  $t(s_{j})$. M is the training dataset that contains all set of possible triplets. $T(s_i, s_i^{+}, s_i^{-})$ is the triplet loss function. This is decomposed into two terms, one that brings the supporting sample closer and one that pushes the contrasting sample further. This is given by 
\begin{equation} 
    T(s_i,s_i^+,s_i^-) =\texttt{max}(0,  D^{+} + \alpha - D^{-})
    \label{triplet_loss}
\end{equation}
\noindent Here $D^+,D^-$ represent the euclidean distance between the target and supporting sample, and target and opposing sample respectively. The parameter $\alpha(=0.2)$ controls the separation margin between these % supporting and contrasting exemplars 
and is obtained through validation data.
% \begin{dmath}\label{triplet_loss}
% T(s_i,s_i^+,s_i^-) = \texttt{max}(0, ||t(s_{i})-t(s_{i}^{+})||^{2}_{2} + \alpha - ||t(s_{i})-t(s_{i}^{-})||^{2}_{2})
% \end{dmath}
% % small t 
% \begin{equation}
% % \scriptsize
% T(s_i,s_i^+,s_i^-) = \texttt{max}(0, ||t(s_{i})-t(s_{i}^{+})||^{2}_{2} + \alpha - ||t(s_{i})-t(s_{i}^{-})||^{2}_{2})
% % \end{dmath}
% \label{triplet_loss}
% \end{equation}
% \begin{equation}%\label{triplet_loss}
%     \begin{split}
% &T(s_i,s_i^+,s_i^-) \\ 
% &=\texttt{max}(0, ||t(s_{i})-t(s_{i}^{+})||^{2}_{2} 
%  + \alpha - ||t(s_{i})-t(s_{i}^{-})||^{2}_{2})
%     \end{split}
%     \label{triplet_loss}
% \end{equation}
% \begin{equation}%\label{triplet_loss}
% T(s_i,s_i^+,s_i^-) =\texttt{max}(0,  D^{+} + \alpha - D^{-})
%     \label{triplet_loss}
% \end{equation}
% The constants $\nu$ and $\alpha$ are obtained through validation data.
%This helps in yielding an attention mechanism that is closer to the supporting example than the contrasting example. The sharing of parameters ensures that the attention mechanism is learnt for both, the visual question answering task as well as the task to obtain a triplet based attention mechanism. 
% We further extend the model to a quadruplet setting where we bring two supporting exemplar embeddings closer and push two contrasting exemplar embeddings far away from the target embedding in a metric learning setting. We provide detailed analysis of this network in the results section in supplementary material.

\subsection{Decoder: Question Generator}
%%%%%%%%%%%%%%%%%%%%%%%%%%%%%%%%%%%%%%%%%%%%%%%%%%%%%%%%%%%%%%%%%%%%%%%
%%%%%%%%%%%%%%%%%%%%%%%%%%%%%%%%%%%%%%%%%%%%%%%%%%%%%%%%%%%%%%%%%%%%%%%
\begin{figure*}[ht]
%\vspace{1in}
	\centering
	\includegraphics[width=0.8 \textwidth]{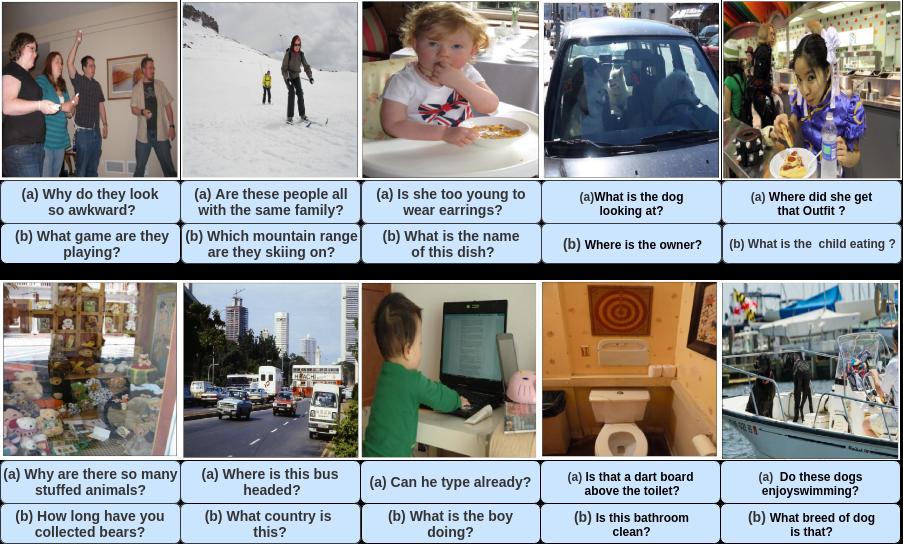}
	\vspace{-0.2cm}
	\caption{These are some examples from the VQG-COCO dataset which provide a comparison between our generated questions and human annotated questions. (a) is the human annotated question for all the images. More qualitative results are present in the supplementary material.}
	\label{fig:natural2}
\end{figure*}
\noindent The role of decoder is to predict the probability for a question, given $s_i$. RNN provides a nice way to perform conditioning on  previous state value using a fixed length hidden vector. 
The conditional probability of a question token at particular time step ${q_{t}}$ is modeled using an LSTM as used in machine translation~\cite{Sutskever_NIPS2014}. 
At time step $t$, the conditional probability is denoted by $P( {q_{t}} | {I ,C},{q_0},...{q_{t-1}})= P( {q_{t}} | {I ,C},h_{t})$, where $h_{t}$ is the hidden state of the LSTM cell at time step $t$, which is conditioned on all the previously generated words $\{{q_0},{q_1},...{q_{N-1}}\}$.
% The newly generated word $q_{t}$.
%$h_{t}$ can be expressed as a linear combination of $h_{t-1}$ and $q_{t}$, $h_{t}= \tanh{(W_h*h_{t-1} + W_e * q_{t})}$. where $h_{t-1}$ is the hidden state at time step $t-1$. 
The word with maximum probability in the probability distribution of the LSTM cell at step $k$  is fed as an input to the LSTM cell at step $k+1$ as shown in part 3 of Figure~\ref{fig:MDN}. At $t=-1$, we are feeding the output of the mixture module to LSTM. $\hat{Q}=\{\hat{q_0},\hat{q_1},...\hat{q_{N-1}}\}$ are the predicted question tokens for the input image $I$. Here, we are using $\hat{q_0}$ and $\hat{q_{N-1}}$ as the special token START and STOP respectively. 
% The question words are fed to LSTM cell at different time steps.
The softmax probability for the predicted question token at different time steps is given by the following equations where LSTM refers to the standard LSTM cell equations:
\begin{equation*}
 \begin{split}
& x_{-1}=S_i=\mbox{Mixture Module}(g_{i},f_{i}) \\
& h_0=\mbox{LSTM}(x_{-1})\\
& x_t=W_e*q_t,  \forall t\in \{0,1,2,...N-1\} \\
& {h_{t+1}}=\mbox{LSTM}(x_t,h_{t}), \forall t\in \{0,1,2,...N-1\}\\
& o_{t+1} = W_o * h_{t+1} \\
& \hat{y}_{t+1} = P( q_{t+1} | {I ,C},h_{t})= \mbox{Softmax}(o_{t+1})\\
& Loss_{t+1}=loss(\hat{y}_{t+1},y_{t+1})
 \end{split}
\end{equation*}
 Where  $\hat{y}_{t+1}$ is the probability distribution over all question tokens. $loss$ is cross entropy loss.
% \begin{equation}
%  \hat{Q}=\arg\max_{ Q^{'}}\ { log\ P( Q^{'} |\ I ,C:\theta )} 
% \end{equation}

% The LSTM model is trained to predict each word in the  question sentence after receiving image and context content from the mixture model and all the previous question conditioning probabilities ($P( q_t | {I ,C},q_0,..,q_{t-1})$) as the initial input to the LSTM cell. Image,context and question words are mapped to common space using CNN, context net and word embedding $W_e$ respectively.
% Here, we are using $q_0$ and $q_N$ as the special token START and STOP respectively.  

% Grammarly after this is left
\subsection{Cost function}
\noindent Our objective is to minimize the total loss, that is the sum of cross entropy loss and triplet loss over all training examples.
  The total loss is: 
  \begin{equation}
%  L( \textbf{s},\textbf{y},\theta)
    L= \frac{1}{M} \sum^{M}_{i=1} (L_{cross} + \gamma L_{triplet})
\end{equation}
where  $M$ is the total number of samples,$\gamma$ is a constant, which controls both the loss. $L_{triplet}$ is the triplet loss function~\ref{triplet_loss}. $L_{cross}$ is the cross entropy loss between the predicted and ground truth questions and is given by: 
 \begin{equation*}
L_{cross}=\frac{-1}{N}\sum_{t=1}^{N} {y_{t} \texttt{log} P(\hat{q_{t}}|I_i,C_i,{\hat{q_0},..\hat{q_{t-1}}})}
\end{equation*}
where, $N$ is the total number of question tokens, $y_t$ is the ground truth label. The code for  MDN-VQG model is provided \footnote{The github link for MDN-VQG Model is \url{https://github.com/MDN-VQG/EMNLP-MDN-VQG}}. 

%
%%%%%%%%%%%%%%%%%%%%%%%%%%%%%%%%%%%%%%%%%%%%%%%%%%%%%%%%%%%%%%%%%%%%%%%%%%%%%%%%%%%%%%%%%%%%%%%%%%%%%%%%%%

\subsection{Variations of Proposed Method}
\label{subsec:variants}
%%%%%%%%%%%%%%%%%%%%%%%%%%%%%%%%%%%%%%%%%%%%%%%%%%%%%%%%%%%%%%%%%%%%%%%
While, we advocate the use of multimodal differential network for generating embeddings that can be used by the decoder for generating questions, we also evaluate several variants of this architecture. These are as follows:

%As human brain generates questions by using various types of information related to an image to generate questions like what is there in the picture and where is this picture taken. So different types of contexts will generate different question. 
% Here we have analyzed with  other context  types of context Place, Tag from captions, nearest objects  in exemplar Image.

 %We propose a triplet network similar to MDN for obtaining the exemplar caption based context embedding, . But the main difference is the  target network consists of CNN only for image feature extraction rather than both image and caption encoding.
\textbf{Tag Net}: In this variant, we consider extracting the part-of-speech (POS) tags for the words present in the caption and obtaining a Tag embedding by considering different methods of combining the one-hot vectors. Further details and experimental results are present in the supplementary.
% clustered average word embeddings for the different tags. 
This Tag embedding is then combined with the image embedding and provided to the decoder network.  

\textbf{Place Net}: In this variant we explore obtaining embeddings based on the visual scene understanding. This is obtained using a pre-trained PlaceCNN~\cite{Zhou_PAMI2017} that is trained to classify 365 different types of scene categories. We then combine the activation map for the input image and the VGG-19 based place embedding to obtain the joint embedding used by the decoder. 

% and is a combination using these two embeddings.%Instead of caption, we Visual object and scene recognition plays a crucial role in the image. Here, places in the image are  labeled with scene semantic categories\cite{Zhou_PAMI2017}, comprise of large and diverse type of environment in the world , such as (park, tower, swimming pool, rain-forest, etc.). So we have used different type of scene semantic categories present in the image as a place based context to generate natural question. A place365 is a convolution neural network is modeled to classify 365 types of scene categories, which is trained on the place2 dataset consist of 1.8 million of scene images.
\textbf{Differential Image Network}: Instead of using multimodal differential network for generating embeddings, we also evaluate differential image network for the same. In this case, the embedding does not include the caption but is based only on the image feature. We also experimented with using multiple exemplars and random exemplars.

\noindent Further details, pseudocode and results regarding these are present in the supplementary material.
%Its is based on the part-of-speech(POS) tag present in the caption. The tags are clustered into three category such as noun tag, verb tag and question tags (What, Where, ...). Joint encoding vector is obtained by applying  word embedding on tag-word followed by temporal convolutional neural network followed by max-pooling network.

%\textbf{Exemplar Image Net }:We propose a triplet network similar to MDN for obtaining the exemplar caption based context embedding, . But the main difference is the  target network consists of CNN only for image feature extraction rather than both image and caption encoding.

% We provide results for all these variations of semantic embeddings and observe that the proposed multimodal differential network provides improved embeddings to the decoder. We believe that these embeddings capture instance specific differential information that helps in disambiguating the kind of question that can be asked for each image. These are further explained in the supplementary material.
% \vspace{-1.3cm}
\subsection{Dataset}
% \vspace{-0.7cm}
We conduct our experiments on  Visual Question Generation (VQG) dataset~\cite{mostafazadeh2016generating}, which contains human annotated questions based on images of MS-COCO dataset. %Visual question generation data set consist of three datasets, VQG-COCO, VQG-Bing, and VQG-Flickr.
This dataset was developed for generating natural and engaging questions based on common sense reasoning. We use VQG-COCO dataset for our experiments which contains a total of 2500 training images, 1250 validation images, and 1250 testing images. Each image in the dataset contains five natural questions and five ground truth captions. It is worth noting that the work of~\cite{jain2017creativity} also used the questions from VQA dataset~\cite{VQA} for training purpose, whereas the work by~\cite{mostafazadeh2016generating} uses only the VQG-COCO dataset.
VQA-1.0 dataset is also built on images from MS-COCO dataset. It contains a total of 82783 images for training, 40504 for validation and 81434 for testing. Each image is associated with 3 questions. We used pretrained caption generation model \cite{Karpathy_CVPR2015} to extract captions for VQA dataset as the human annotated captions are not there in the dataset. We also get good results on the VQA dataset (as shown in Table~\ref{score_tab_2}) which shows that our method doesn't necessitate the presence of ground truth captions. We train our model separately for VQG-COCO and VQA dataset.
% the generalizability of our method to generated captions 

\subsection{Inference}
We made use of the 1250 validation images to tune the hyperparameters and are providing the results on test set of VQG-COCO dataset. 
% Since, the triplet model is learned to bring supporting feature close to target feature and constricting feature far away from the target feature. 
During inference, We use the Representation module to find the embeddings for the image and ground truth caption without using the supporting and contrasting exemplars. The mixture module provides the joint representation of the target image and ground truth caption. Finally, the decoder takes in the joint features and generates the question. We also experimented with the captions generated by an Image-Captioning network~\cite{Karpathy_CVPR2015} for VQG-COCO dataset and the result for that and training details are present in the supplementary material.
% Training details are given in the supplementary material.

\begin{figure}[ht]
	\centering
	\includegraphics[width=0.4\textwidth]{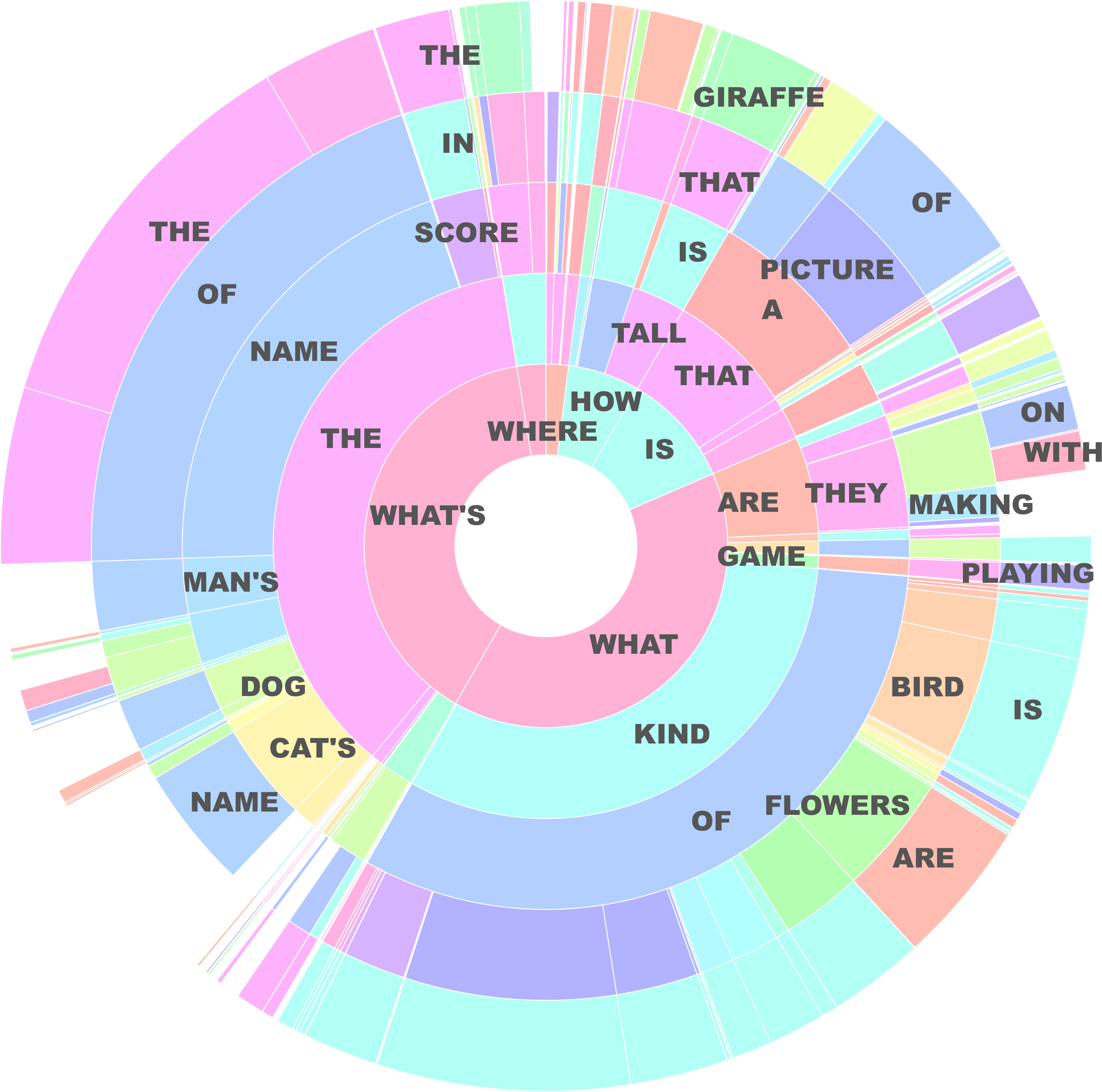}
% 	\vspace{-0.7cm}
	\caption{Sunburst plot for VQG-COCO: The $i^{th}$ ring captures the frequency distribution over words for the $i^{th}$ word of the generated question. The angle subtended at the center is proportional to the frequency of the word. While some words have high frequency, the outer rings illustrate a fine blend of words. We have restricted the plot to 5 rings for easy readability. Best viewed in color.}
	\label{tbl:sunburst}
\end{figure}

\section{Experiments}
We evaluate our proposed MDN method in the following ways: First, we evaluate it against other variants described in section~\ref{subsec:variants} and ~\ref{mixture_model}. Second, we further compare our network with state-of-the-art methods  for VQA 1.0 and VQG-COCO dataset.
% i.e.~\cite{Yang_arXiv2015} for VQA 1.0 
% % and Natural~\cite{Mostafazadeh_ACL2016} 
% and Creative~\cite{jain_CVPR2017} for VQG-COCO dataset.
 We perform a user study to gauge human opinion on naturalness of the generated question and analyze the word statistics in Figure~\ref{tbl:sunburst}. This is an important test as humans are the best deciders of naturalness. We further consider the statistical significance for the various ablations as well as the state-of-the-art models. The quantitative evaluation is conducted using standard metrics like BLEU~\cite{Papineni_ACL2002}, METEOR~\cite{Banerjee_ACL2005}, ROUGE~\cite{Lin_ACL2004}, CIDEr~\cite{Vedantam_CVPR2015}. Although these metrics have not been shown to correlate with `naturalness' of the question these still provide a reasonable quantitative measure for comparison. %Further details about these metrics are present in the supplementary material. 
Here we only provide the BLEU1 scores, but the remaining BLEU-n metric scores are present in the supplementary. We observe that the proposed MDN provides improved embeddings to the decoder. We believe that these embeddings capture instance specific differential information that helps in guiding the question generation. Details regarding the metrics  are given in the supplementary material.
% These are further explained in the supplementary material.

%%%%%%%%%%%%%%%%%%%%%%%%%%%%%%%%%%%%%%%%%%%%%%%%%%%%%%%%%%%%%%%%%%%%%%%
\label{sec:results}
% %%%%%%%%%%%%%%%%%%%%%%%%%%%%%%%%%%%%%%%%%%%%%%%%%%%%%%%%%%%%%%%%%%%%%%%   

% %fig:natural4
 %%%%%%%%%%%%%%%%%%%%%%%%%%%%%%%%%%%%%%%%%%%%%%%%%%%%%%%%%%%%%%%%%%%%%%% 
\subsection{Ablation Analysis}\label{ablation_analysis}
% We performed ablation analysis for our proposed method in which 
We considered different variations of our method mentioned  in section~\ref{subsec:variants} and the various ways to obtain the joint multimodal embedding as described in section~\ref{mixture_model}. The results for the VQG-COCO test set are given in table~\ref{score_tab_1}. In this table, every block provides the results for one of the variations of obtaining the embeddings and different ways of combining them. We observe that the Joint Method (JM) of combining the embeddings works the best in all cases except the Tag Embeddings. Among the ablations, the proposed MDN method works way better than the other variants in terms of BLEU, METEOR and ROUGE metrics by achieving an improvement of 6\%, 12\% and 18\% in the scores respectively over the best other variant. 
% The METEOR metric is considered closest to human evaluation~\cite{Banerjee_ACL2005}.
% The generated questions for these different variants are shown in figure~\ref{fig:examples_ablations}.
%\subsubsection{Ablation Analysis of Network with contexts}
\begin{table}[ht]
\scriptsize
\centering
\begin{tabular}{|l|l|cccc|}
\hline \bf Emb. & \bf Method & \bf BLEU1 & \bf METEOR & \bf ROUGE & \bf CIDEr \\ \hline 

Tag & AtM & 22.4 &8.6 & 22.5 & 20.8\\ 
Tag & HM & \textbf{24.4} &\textbf{10.8}& \textbf{24.3} & \textbf{55.0}\\
% Tag & HM &0.248&0.106& 0.244 & \bf0.532\\
Tag & AM & 24.4 &10.6& 23.9 & 49.4\\
Tag& JM &22.2 &10.5 & 22.8 &50.1 \\ \hline

PlaceCNN & AtM &24.4  &10.3 &24.0 &51.8 \\
PlaceCNN & HM &24.0  &10.4 &24.3 &49.8 \\
PlaceCNN & AM & 24.1 & 10.6&24.3 &51.5 \\
PlaceCNN & JM &\textbf{25.7}  &\textbf{10.8 }&\textbf{24.5} &\textbf{56.1}  \\ \hline

Diff. Img & AtM & 20.5 &8.5 & \textbf{24.4} & 19.2\\ 
Diff. Img & HM& 23.6 &8.6 & 22.3 & 22.0\\
Diff. Img & AM & 20.6 &8.5 & 24.4 & 19.2\\ 
Diff. Img & JM & \textbf{30.4} & \textbf{11.7} & 22.3 & \textbf{22.8}\\ \hline

MDN & AtM & 22.4 &8.8 & 24.6 & 22.4\\ 
MDN & HM & 26.6 &12.8 & 30.1 & 31.4\\
MDN & AM & 29.6 &15.4 & 32.8 & 41.6\\ 
MDN \textbf{(Ours)} & JM & \textbf{36.0}&\textbf{23.4}&\textbf{41.8}& \textbf{50.7}\\\hline

\end{tabular}
% \vspace{-0.4cm}
\caption{\label{score_tab_1}Analysis of variants of our proposed method on VQG-COCO Dataset as mentioned in section~\ref{subsec:variants} and different ways of getting a joint embedding (Attention (AtM), Hadamard (HM), Addition (AM) and Joint (JM) method as given in section~\ref{mixture_model}) for each method. Refer section~\ref{ablation_analysis} for more details.
% The first, second, third and fourth  blocks provide results for Tag-Net, Place-Net, Diff. Image Net and our MDN method respectively.
}
\vspace{-0.5cm}
\end{table}
%%%%%%%%%%%%%%%%%%%%%%%%%%%%%%%%%%%%%%%%%%%%%%%%%%%%%%%%%%%%%%%%%%%%%%%
\subsection{Baseline and State-of-the-Art}\label{baseline_sota}
% We obtain the comparison with the baselines against the questions that are provided by  human consensus for the VQG task. 
The comparison of our method with various baselines and state-of-the-art methods is provided in table~\ref{score_tab_2} for VQA 1.0 and table~\ref{score_tab_3} for VQG-COCO dataset. The comparable baselines for our method are the image based and caption based models in which we use either only the image or the caption embedding and generate the question. In both the tables, the first block consists of the current state-of-the-art methods on that dataset and the second  contains the baselines. We observe that for the VQA dataset we achieve an improvement of 8\% in BLEU and 7\% in METEOR metric scores over the baselines, whereas for VQG-COCO dataset this is 15\% for both the metrics. We improve over the previous state-of-the-art~\cite{Yang_arXiv2015} for VQA dataset by around 6\% in BLEU score and 10\% in METEOR score. In the VQG-COCO dataset, we improve over~\cite{mostafazadeh2016generating} by 3.7\% and~\cite{jain2017creativity} by 3.5\% in terms of METEOR scores.

\begin{table}[ht]
\scriptsize
\centering
\begin{tabular}{|l|cccc|}
\hline \bf Methods & \bf BLEU1  & \bf METEOR & \bf ROUGE &  \bf CIDEr \\ \hline
Sample(Yang,\citeyear{Yang_arXiv2015}) & 38.8  & 12.7 & 34.2 & 13.3 \\
Max(Yang,\citeyear{Yang_arXiv2015}) &  59.4  & 17.8 & 49.3 & 33.1\\ \hline
Image Only & 56.6  & 15.1  & 40.0 & 31.0\\
Caption Only & 57.1  & 15.5  & 36.6 & 30.5\\ \hline
MDN-Attention   & 60.7  & 16.7  & 49.8 & 33.6\\
MDN-Hadamard  & 61.7 & 16.7  & 50.1 & 29.3 \\
MDN-Addition  & 61.7  & 18.3  & 50.4 & \textbf{42.6} \\
MDN-Joint (\bf Ours)& \textbf{65.1}  & \textbf{22.7}  & \textbf{52.0} & 33.1\\
\hline
\end{tabular}
\caption{\label{score_tab_2}State-of-the-Art comparison on VQA-1.0 Dataset. The first block consists of the state-of-the-art results, second block refers to the baselines mentioned in section~\ref{baseline_sota}, third block provides the results for the variants of mixture module present in section~\ref{mixture_model}.}
\end{table}
% SOTA comparison on VQA-1.0 Dataset for MDN. 
% \vspace{-0.5cm}
\begin{table}[ht]
\scriptsize
\centering
\begin{tabular}{|l|lccc|}
\hline \bf Context &  \bf BLEU1 & \bf METEOR & \bf ROUGE & \bf CIDEr \\ \hline 

Natural\citeyear{mostafazadeh2016generating} & 19.2 & 19.7  &- & -  \\
Creative\citeyear{jain2017creativity} & {35.6} & 19.9 & - & - \\ \hline
Image Only &  20.8  &  8.6  & 22.6 & 18.8\\
Caption Only & 21.1 & 8.5  & 25.9 & 22.3\\\hline
Tag-Hadamard & {24.4} &{10.8}& {24.3} & {55.0}\\ 
PlaceCNN-Joint & {25.7}  &{10.8}&{24.5} &\textbf{56.1} \\
Diff.Image-Joint& 30.4 & {11.7} & 26.3 & {38.8}\\
MDN-Joint (\bf Ours)& \textbf{36.0}&\textbf{23.4}&\textbf{41.8}& 50.7\\\hline
Humans\citeyear{mostafazadeh2016generating} & \textbf{86.0}&\textbf{60.8}&\textbf{-}& -\\\hline
\end{tabular}
\caption{\label{score_tab_3}State-of-the-Art (SOTA) comparison on VQG-COCO Dataset. The first block consists of the SOTA results, second block refers to the baselines mentioned in section~\ref{baseline_sota}, third block shows the results for the best method for different ablations mentioned in table~\ref{score_tab_1}. }
\end{table}

 %%%%%%%%%%%%%%%%%%%%%%%%%%%%%%%%%%%%%%%%%%%%%%%%%%%%%%%%%%%%%%%%%%%%%%% 
\subsection{Statistical Significance Analysis}
\begin{figure}[ht]
	\centering
	\includegraphics[width=0.45\textwidth]{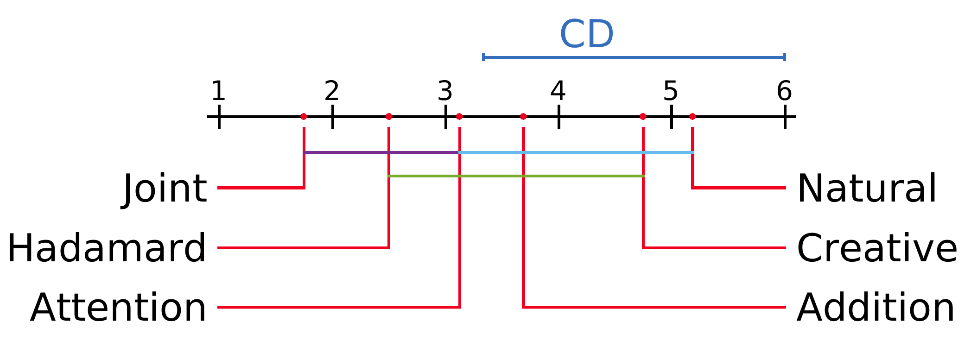}
	\vspace{-0.35cm}
	\caption{The mean rank of all the models on the basis of METEOR score are plotted on the x-axis. Here Joint refers to our MDN-Joint model and others are the different variations described in section~\ref{mixture_model} and Natural~\cite{mostafazadeh2016generating}, Creative~\cite{jain2017creativity}. The colored lines between the two models represents that these models are not significantly different from each other.}
	\label{fig:result_1_B}
\end{figure}

We have analysed Statistical Significance~\cite{Demvsar_JMLR2006} of our MDN model for VQG for different  variations of the mixture module mentioned in section~\ref{mixture_model} and also against the state-of-the-art methods. 
% as well as different methods of fusing the cues. 
The Critical Difference (CD) for Nemenyi~\cite{Fivser_PLOS2016} test depends upon the given $\alpha$ (confidence level, which is 0.05 in our case) for average ranks and N (number of tested datasets). If the difference in the rank of the two methods lies within CD, then they are not significantly different and vice-versa. Figure~\ref{fig:result_1_B} visualizes the post-hoc analysis using the CD diagram. From the figure, it is clear that MDN-Joint works best and is statistically significantly different from the state-of-the-art methods. 
% \vspace{-0.75cm}
\begin{figure}[ht]
	\centering
	\includegraphics[width=0.5\textwidth]{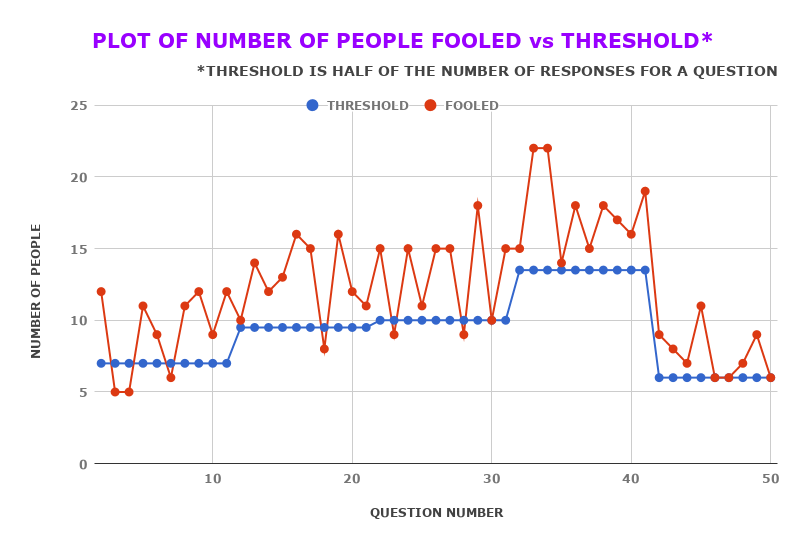}
	\vspace{-0.89cm}
	\caption{Perceptual Realism Plot for human survey. Here every question has different number of responses and hence the threshold which is the half of total responses for each question is varying. This plot is only for 50 of the 100 questions involved in the survey. See section~\ref{percep_real} for more details.}
	\label{fig:result_2_A}
\end{figure}

 %%%%%%%%%%%%%%%%%%%%%%%%%%%%%%%%%%%%%%%%%%%%%%%%%%%%%%%%%%%%%%%%%%%%%%% 
% We can see Joint method is statistically significantly over all of the models.The purple and green lines indicate groups of models which are not significantly different (their average ranks differ by less than CD value)

%  \vspace{-0.5cm}
\subsection{Perceptual Realism}\label{percep_real}
%%%%%%%%%%%%%%%%%%%%%%%
A human is the best judge of naturalness of any question, We evaluated our proposed MDN method using a `Naturalness' Turing test~\cite{Zhang_ECCV2016} on 175 people. 
% We asked human participants to choose the natural question between questions generated by our model and ground truth questions. 
People were shown an image with 2 questions just as in figure~\ref{fig:vqg_result} and were asked to rate the naturalness of both the questions on a scale of 1 to 5  where 1 means `Least Natural' and 5 is the `Most Natural'. We provided 175 people with 100 such images from the VQG-COCO validation dataset which has 1250 images. % and this survey was done on around 175 people.
Figure~\ref{fig:result_2_A} indicates the number of people who were fooled (rated the generated question more or equal to the ground truth question). For the 100 images, on an average 59.7\% people were fooled in this experiment and this shows that our model is able to generate natural questions. 
\section{Conclusion}
In this paper we have proposed a novel method for generating natural questions for an image. The approach relies on obtaining multimodal differential embeddings from image and its caption.
% Our approach has also been analysed in terms of other ablations
% variants such as using differential image, tag and place based combined embeddings
% and we observe that the proposed Multimodal Differential Embeddings performs the best.
We also provide ablation analysis and a detailed comparison with state-of-the-art methods, perform a user study to evaluate the naturalness of our generated questions and also ensure that the results are statistically significant.
In future, we would like to analyse means of obtaining composite embeddings. We also aim to consider the generalisation of this approach to other vision and language tasks.
%%%%%%%%%%%%%%%%%%%%%%%%%%%%%%%%%%%%%%%%%%%%%%%%%%%%%%%%%%%%%%%%%%%%%%%
\bibliography{emnlp2018}
\bibliographystyle{acl_natbib_nourl}
%%%%%%%%%%%%%%%%%%%%%%%%%%%%%%%%%%%%%%%%%%%%%%%%%%%%%%%%%%%%%%%%%%%%%%%

% \begin{abstract}
%   In the supplementary material, we provide details regarding the experimental setup used while training the proposed method and details about the datasets used. We also provide detailed explanation for variants of the proposed methods for generating natural question based on the image. We further provide additional results for the different variants used. We give the pseudocode for our method and also explain different fusion methods used in the Mixture module. 
% %   We provide additional perceptual realism results for our generated questions. %Finally we provide different score visualizations and training loss for our models.
% \end{abstract}
\appendix
%%%%%%%%%%%%%%%%%%%%%%%%%%%%%%%%%%%%%%%%%%%%%%%%%%%%%%%%%%%%%%%%%%%%%%%%%%%%%%%%%%
\section{Supplementary Material}
%%%%%%%%%%%%%%%%%%%%%%%%%%%%%%%%%%%%%%%%%%%%%%%%%%%%%%%%%%%%%%%%%%%%%%%%%%%%%%%%%%
% In order to understand the progress towards multimodal vision and language understanding, 
Section~\ref{sec-3} will provide details about training configuration for MDN,
Section~\ref{sec-4} will explain the various Proposed Methods and we also provide a discussion in section~\ref{disc} regarding some important questions related to our method. 
% experiment performed on various Dataset and evaluation metric 
We report BLEU1, BLEU2, BLEU3, BLEU4, METEOR, ROUGE and CIDER metric scores for VQG-COCO dataset. We present different experiments with Tag Net in which we explore the performance of various tags (Noun, Verb, and Question tags) and different ways of combining them to get the context vectors.
	\begin{algorithm}
	\caption{Multimodal Differential Network}\label{MC-BMN}
	\begin{algorithmic}[1]
		\Procedure{MDN}{$x_{i}$}
		\BState\emph{ Finding Exemplars}:
		\State ${x_i^+,x_i^-}: =KD-Tree(x_{i})$
		\State ${c_i,c_i^+,c_i^- :=} Extract\_caption(x_{i},x_i^+,x_i^-)$
		\BState\emph{ Compute Triplet Embedding}:
		\State  ${g_i,g_i^+,g_i^-}:=Triplet\_CNN(x_{i},x_i^+,x_i^-)$ 
		\State  ${f_i,f_i^+,f_i^-:=}Triplet\_LSTM(c_{i},c_i^+,c_i^-)$

		\BState\emph{ Compute Triplet Fusion Embedding }:
% 		\State ${s_i,s_i^+,s_i^-}:=Triplet\_fuse({g_i,g_i^+,g_i^-},{f_i,f_i^+,f_i^-})$
			\State $s_i=Triplet\_Fusion(g_{i},f_{i}, Joint) $
			\State $s^+_i=Triplet\_Fusion(g_{s},f_{s}, Joint)$
			\State $s^-_i=Triplet\_Fusion(g_{c},f_{c}, Joint)$
		\BState\emph{ Compute Triplet Loss}:
		 \State $Loss\_Triplet= triplet\_loss(s_i,s^+_i,s^-_i)$

% 		\BState \emph{Compute Fusion Distribution}:
% 		\State  Place Fusion: $\mu_{P}=BayesianFuse(g_p,g_i)$
% 		\State  Caption Fusion: $\mu_{C}=BayesianFuse(g_c,g_i)$
% 		\State  Tag Fusion: $\mu_{T}=BayesianFuse(g_t,g_i)$

		\BState \emph{Compute Decode Question Sentence}:
		\State   $\hat{y}=Generating\_LSTM(s_i,h_i,c_i)$
		\State   $loss=Cross\_Entropy(y,\hat{y})$
		\EndProcedure
	\State {-----------------------------------------------------}
	\Procedure{Triplet Fusion}{$g_{i}$,$f_{i},flag$}
			\State $g_{i}$:Image feature,14x14x512
			\State $f_{i}$: Caption feature,1x512
				\BState \emph{Match Dimension}:
			\State $G_{img}=reshape(g_{i})$,196x512
			\State  $F_{caps}=clone(f_{i})$ 196x512
% 			\BState \emph{Match Dimension}:
% 			\State $G_{imgfeat}$:Reshape  $g_{i}$ to 196x512 :$reshape(g_{i})$
% 			\State  $F_{quesfeat}$: Replicate $f_i$ to 196 times: $clone(f_{i})$ 
% 			\BState \emph{If flag==Hadamard Fusion}:
%             \State $A_{had}=   \tanh( W_{ih}  G_{img}    \emul \ ( W_{ch} F_{cap} + b_h))$
% 			\State $S_{emb}=\tanh({W_A} A_{had} + b_A)$
			
% 			\BState \emph{If flag==Addition Fusion }:
%             \State $A_{add}=   \tanh( W_{ia}  G_{img}  \eadd ( W_{ca}  F_{cap}  + b_a))$
% 			\State $S_{emb}=\tanh({W_A} A_{add} + b_A)$, 
			
			\BState \emph{If flag==Joint Fusion}:
            \State $A{jnt}=   \tanh(  W_{ij}  G_{img} \boxdot (W_{cj}  F_{cap} + b_j))$
			\State $S_{emb}=\tanh({W_A} A_{jnt} + b_A)$, 
			\State [$\boxdot = *$  (MDN-Mul), $\boxdot = +$  (MDN-Add)]
			
			\BState \emph{If flag==Attention Fusion }:
			\State $h_{att}= \tanh({W_I}{G_{img}} \eadd ({W_C}{F_{cap}}+{b_c}))$
			\State $P_{att}= \mbox{Softmax}({W_P}{h_{att}}+{b_P})$
			\State $V_{att}= \sum_{i}{P_{att}(i)}{G_{img}(i)}$
			\State $A_{att}= V_{att} + f_{i}$
			\State $S_{emb}=\tanh({W_A} A_{att} + b_A)$
			
			\BState \emph{ Return $S_{emb}$}
			\EndProcedure
	\end{algorithmic}
\end{algorithm}

% In section~\ref{sec-4}, we present additional quantitative results.  
%%%%%%%%%%%%%%%%%%%%%%%%%%%%%%%%%%%%%%%%%%%%%%%%%%%%%%%%%%%%%%%%%%%%%%%%%%%%%%

\begin{figure*}[ht]
	\includegraphics[width=0.9 \textwidth]{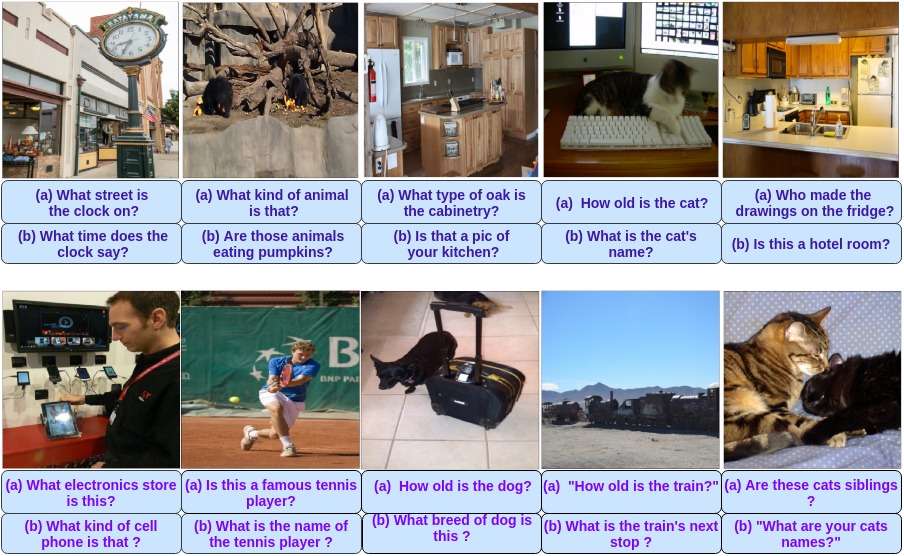}
	\caption{These are some more examples from the VQG-COCO dataset which provide a comparison between the questions generated by our model and human annotated questions. (b) is the human annotated question for the first row-fourth column, \& fifth column image and (a) for the rest of images.}
	\label{fig:natural}
\end{figure*}

% %-------------------------------------------------------------------------

% \subsection{Analysis of Context: Tag }
% \label{sec:context_analysis_tag}
\begin{table}[ht]
\scriptsize
\centering
\begin{tabular}{|l|l|cccc|}
\hline \bf Context & \bf Meth & \bf BLEU-1 & \bf Meteor & \bf Rouge & \bf CIDer \\ \hline 
Image & - & 23.2  &  8.6  & 25.6 & 18.8\\
Caption  &- & 23.5 & 8.6  & 25.9 & 24.3\\\hline
Tag-n & JM &22.2 &10.5 & 22.8 &50.1 \\
Tag-n & AtM & 22.4 &8.6 & 22.5 & 20.8\\
Tag-n & HM &\bf24.8&\bf10.6& 24.4 & \bf53.2\\
Tag-n & AM &24.4 &10.6& 23.9 & 49.4\\ \hline

Tag-v & JM & 23.9 &10.5  &24.1 &\bf52.9 \\
Tag-v & AtM & 22.2 &8.6 &22.4 &20.9\\
Tag-v & HM &\bf24.5 &\bf10.7&24.2 &\bf52.3 \\
Tag-v & AM & 24.6 &10.6 &24.1 &49.0\\ \hline

Tag-wh & JM &22.4  &10.5 &22.5 &48.6\\
Tag-wh & AtM &22.2  &8.6 &22.4 &20.9\\
Tag-wh & HM & \bf24.6 &\bf10.8& 24.3 & \textbf{55.0}\\
Tag-wh & AM &24.0  &10.4 &23.7 &47.8\\ \hline

\end{tabular}
\caption{\label{score_tab_7}  Analysis of different Tags for VQG-COCO-dataset. We analyse noun tag (Tag-n), verb tag (Tag-v) and question tag (Tag-wh) for different fusion methods namely joint, attention, Hadamard and addition based fusion.}
\end{table}

\begin{table}[ht]
\scriptsize
\centering
\begin{tabular}{|l|l|c  c c|}
\hline \bf Context & \bf BLEU-1 & \bf Meteor & \bf Rouge & \bf CIDer \\ \hline

Tag-n3-add & 22.4 &9.1& 22.2 &26.7\\
Tag-n3-con  &\textbf{24.8}&10.6& 24.4 & \bf53.2\\
Tag-n3-joint& 22.1&8.9&21.7 & 24.6\\
Tag-n3-conv  &24.1&10.3&24.0&47.9\\\hline

Tag-v3-add  &24.1 &10.2& 23.9  &46.7 \\
Tag-v3-con  &24.5 &10.7&24.2 &\bf52.3 \\
Tag-v3-joint  & 22.5&9.1&22.1& 25.6\\
Tag-v3-conv  & 23.2&9.0&24.2 &38.0 \\\hline

Tag-q3-add  & 24.5 &10.5& 24.4 &{51.4}\\
Tag-q3-con  & 24.6 &\textbf{10.8}& 24.3 & \textbf{55.0}\\
Tag-q3-joint  & 22.1 &9.0& 22.0 &25.9\\
Tag-q3-conv  &24.3 &10.4&24.0 &48.6\\\hline
\end{tabular}
\caption{\label{score_tab_9}Combination of 3 tags of each category for hadamard mixture model namely addition, concatenation, multiplication and 1d-convolution}
\end{table}

% \caption{\label{score_tab_9}Analysis of combinations of various Tags on VQG-COCO dataset. Here we take 3 tags of each category and apply different methods to combine them namely addition, concatenation, multiplication and 1d-convolution for Hadamard Model

% %-------------------------------------------------------------------------
% So in this case we use the Differential Image Network for Various models of COCO VQG and found out that K=3 works best in terms of METEOR and BLEU Scores

\begin{table}[ht]
\scriptsize
\centering
\begin{tabular}{|l|c|cccc|}
\hline \bf Meth & \bf Exemplar & \bf BLEU-1 & \bf Meteor & \bf Rouge  & \bf CIDer \\ \hline
AM &IE(K=1)& 21.8 &7.6 & 22.8 & 22.0\\ 
AM &IE(K=2)& 22.4 &8.3 & 23.4 & 16.0\\ 
AM &IE(K=3)& 22.1 &8.8 & 24.7 & 24.1\\ 
AM &IE(K=4)& 23.7 &9.5 &\textbf{ 25.9} & 25.2\\ 
AM &IE(K=5)&\textbf{24.4}  & \textbf{11.7} & 25.0 & {27.8}\\
AM &IE(K=R)& 18.8 & 6.4  & 20.0 & 20.1\\\hline 
HM &IE(K=1)& 23.6 &7.2 & 25.3 & 21.0\\ 
HM &IE(K=2)& 23.2 &8.9 & \textbf{27.8} & 22.1\\ 
HM &IE(K=3)& 24.8 &9.8 & 27.9 & 28.5\\ 
HM &IE(K=4) & 27.7 &9.4 & 26.1 & \textbf{33.8}\\ 
HM &IE(K=5)& \textbf{28.3} &\textbf{10.2}  & 26.6 & 31.5\\
HM &IE(K=R)& 20.1 & 7.7  & 20.1 & 20.5\\\hline 
JM &IE(K=1)& 20.1 &7.9 & 21.8 & 20.9\\ 
JM &IE(K=2)& 22.6 &8.5 & 22.4 & 28.2\\\
JM &IE(K=3)  & 24.0 &9.2 & 24.4 & 29.5\\
JM &IE(K=4)& 28.7 &10.2 & 24.4 & 32.8\\ 
JM &IE(K=5) &\textbf{30.4}  & \textbf{11.7} & \textbf{26.3} & {38.8}\\ 
JM &IE(K=R)& 21.8 & 7.4  & 22.1 & 22.5\\\hline 
\end{tabular}
\caption{\label{score_tab_11}VQG-COCO-dataset, Analysis of different number of Exemplars for addition model, hadamard model and joint model, R is random exemplar. All these experiment are for the differential image network. k=5 performs the best and hence we use this value for the results in main paper.}
\end{table}

\begin{table*}[ht]
% \scriptsize
\centering
\begin{tabular}{|l|lcccccc|}
\hline \bf Context &  \bf BLEU1 & \bf BLEU2  & \bf BLEU3  & \bf BLEU4& \bf METEOR & \bf ROUGE & \bf CIDEr \\ \hline 

Natural \citeyear{mostafazadeh2016generating} & 19.2  & -&- & -  & 19.7  &- & -  \\
Creative \citeyear{jain2017creativity} & {35.6}  & -& -& -&   19.9 & - & - \\ \hline
Image Only &  20.8  & 14.1&8.5 &5.2 &     8.6  & 22.6 & 18.8\\
Caption Only & 21.1  & 14.2& 8.6&5.4 &   8.5  & 25.9 & 22.3\\\hline
Tag-Hadamard & {24.4}  &15.1 & 9.5& 6.3&{10.8}& {24.3} & {55.0}\\ 
PlaceCNN-Joint & {25.7}  & 15.7& 9.9& 6.5  &{10.8}&{24.5} &\textbf{56.1} \\
Diff.Image-Joint& 30.4  & 20.1& 14.3& 8.3 & {11.7} & 26.3 & {38.8}\\
MDN-Joint (\bf Ours)& \textbf{36.0}  & 24.9&16.8 & 10.4&\textbf{23.4}&\textbf{41.8}& 50.7\\\hline
Humans \citeyear{mostafazadeh2016generating} & \textbf{86.0} &- &- &- &\textbf{60.8}&\textbf{-}& -\\\hline
\end{tabular}
\caption{\label{score_tab_3a}Full State-of-the-Art comparison on VQG-COCO Dataset. The first block consists of the state-of-the-art results, second block refers to the baselines mentioned in State-of-the-art section of main paper and the third block provides the results for the best method for different ablations of our method. }
\end{table*}

\begin{figure}[ht]
	\centering
	\includegraphics[width=0.5\textwidth]{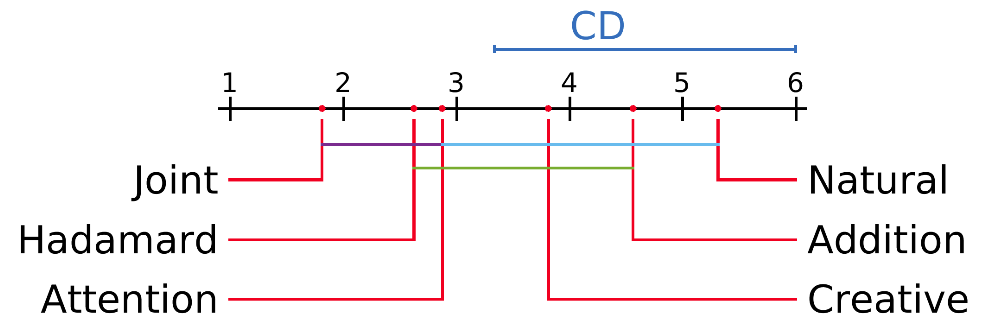}
	\caption{The mean rank of all the models on the basis of BLEU score are plotted on the x-axis. Here Joint refers to our MDN-Joint model and others are the different variations of our model and Natural-\cite{mostafazadeh2016generating}, Creative-\cite{jain2017creativity}. Also the colored lines between two models represent that those models are not significantly different from each other. }
	\label{fig:result_1_A}
\end{figure}
%-------------------------------------------------------------------------
%%%%
\section{Dataset and Training Details}

\label{sec-3}
	\subsection{Dataset}
	We conduct our experiments on two types of dataset: VQA dataset~\cite{VQA}, which contains human annotated questions based on images on MS-COCO dataset. Second one is VQG-COCO dataset based on natural question~\cite{mostafazadeh2016generating}. 
	
	\subsubsection{VQA dataset}
	VQA dataset\cite{VQA} is built on complex images from MS-COCO dataset. It contains a total of 204721 images, out of which 82783 are for training, 40504 for validation and 81434 for testing. Each image in the MS-COCO dataset is associated with 3 questions and each question has 10 possible answers. So there are  248349 QA pair for training, 121512 QA pairs for validating and 244302 QA pairs for testing. We used pre-trained caption generation model \cite{Karpathy_NIPS2014} to extract captions for VQA dataset.

	\subsubsection{VQG dataset}
% 	Visual Question Generation data set consist of three datasets, VQG-COCO, VQG-Bing, and VQG-Flickr.
	The VQG-COCO dataset\cite{mostafazadeh2016generating}, is developed for generating natural and engaging questions that are based on common sense reasoning. This dataset contains a total of 2500 training images, 1250 validation images and 1250 testing images. Each image in the dataset contains 5 natural questions. 
% 	VQG-Bing,VQG-Flickr dataset  pattern is similar to VGQ-COCO.

	\subsection{Training Configuration}
We have used RMSPROP  optimizer to update the model parameter and configured hyper-parameter values to be as follows: {$\text{learning rate}=0.0004, \text{batch size} = 200, \alpha = 0.99, \epsilon=1e-8$} to train the classification network . In order to train a triplet model, we have used RMSPROP to  optimize the triplet model model parameter and configure hyper-parameter values to be: {$\text{learning rate}=0.001, \text{batch size} = 200, \alpha = 0.9, \epsilon=1e-8$}.
% {learning rate =0.001 , batch size = 200, alpha = 0.9 and epsilon=1e-8}. 
We also used learning rate decay to decrease the learning rate on every epoch by a factor given by:
\[Decay\_factor=exp\left(\frac{log(0.1)}{a*b} \right)\] where values of a=1500 and b=1250 are set empirically.

\section{Ablation Analysis of Model}

\label{sec-4}
 While,  we  advocate  the  use  of  multimodal  differential network (MDN) for generating embeddings that can be used by the decoder for generating questions, we also evaluate several variants of this architecture namely (a) Differential Image Network, (b) Tag net and  (c) Place net. These are described in detail as follows:

%%%%%%%%%%%%%%%%%%%%%%%%%%%%%%%%%%%%%%%%%%%%%%%%%%%%%%%%%%%%%%%%%%%%%%%
\subsection{Differential Image Network}
%%%%%%%%%%%%%%%%%%%%%%%%%%%%%%%%%%%%%%%%%%%%%%%%%%%%%%%%%%%%%%%%%%%%%%%

For obtaining the exemplar image based context embedding, we propose a triplet network consist of three network, one is target net, supporting net and opposing net.  All these three networks designed with convolution neural network and shared the same parameters.

\begin{figure}[ht]
	%\vspace{1in}
	\centering
	\includegraphics[width=0.5\textwidth]{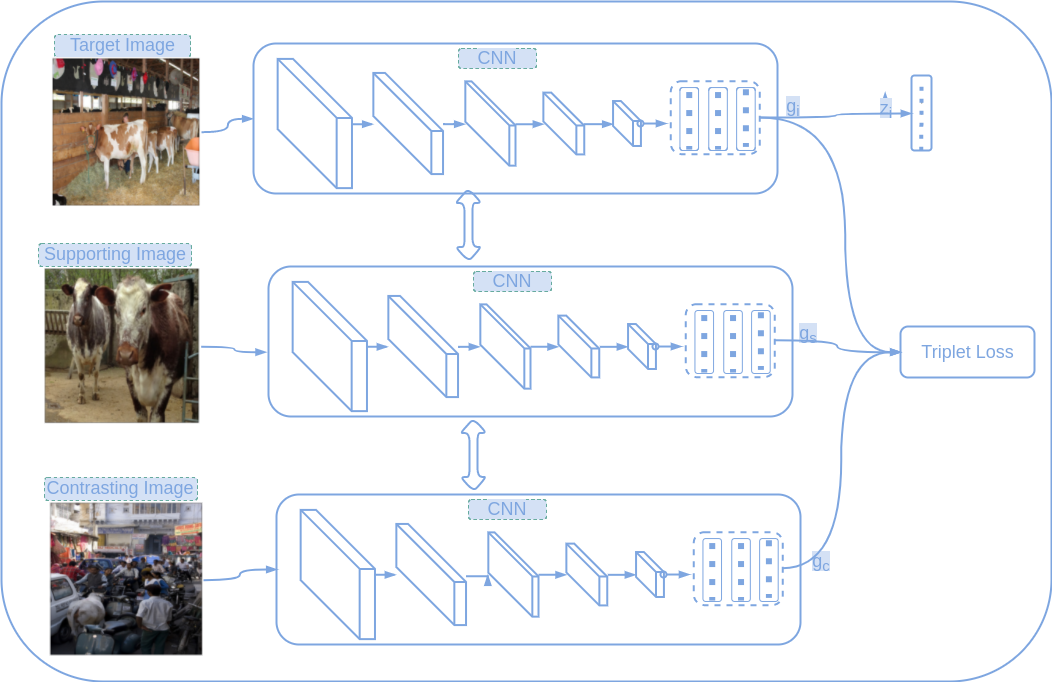}
	\caption{ Differential Image Network }
	\label{fig:DTN}
\end{figure}

The weights of this network are learnt through end-to-end learning using  a triplet loss.  The aim is to obtain latent weight vectors that bring the supporting exemplar close to the target image  and enhances the difference between opposing examples.  More formally, given an image $x_i$ we obtain an embedding $g_i$ using a CNN that we parameterize through a function $G(x_i,W_c)$ where $W_c$ are the weights of the CNN. This is illustrated in  figure~\ref{fig:DTN}.   
% $T(g_i, g_i^{+}, g_i^{-})$ is the triplet loss function that is used. This is decomposed into two terms, one that brings the positive sample closer and one that pushes the negative sample farther. This is given by
%%%%%%%%%%%%%%%%%%%%%%%%

%%%%%%%%%%%%%%%%%%%%%%%%%%%%%%%%%%%%%%%%%%%%%%%%%%%%%%%%%%%%%%%%%%%%%%%%%%%%%%%%%%
% \section{Explanation of other Proposed Methods}
% \label{sec-2}
%%%%%%%%%%%%%%%%%%%%%%%%%%%%%%%%%%%%%%%%%%%%%%%%%%%%%%%%%%%%%%%%%%%%%%%%%%%%%%%%%%

%%%%%%%%%%%%%%%%%%%%%%%%%%%%%%%%%%%%%%%%%%%%%%%%%%%%%%%%%%%%%%%%%%%%%%%
\subsection{Tag net}
\begin{figure}[ht]
	%\vspace{1in}
	\centering
	\includegraphics[width=0.5\textwidth]{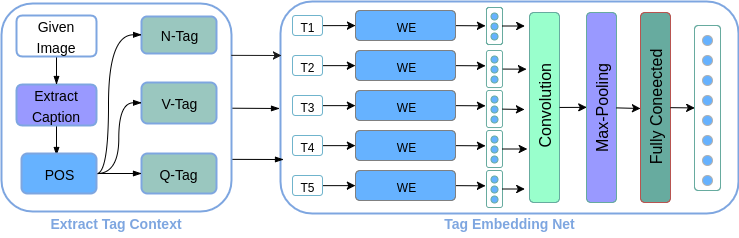}
	\caption{ Illustration of Tag Net }
	\label{fig:tag}
\end{figure}

%%%%%%%%%%%%%%%%%%%%%%%%%%%%%%%%%%%%%%%%%%%%%%%%%%%%%%%%%%%%%%%%%%%%%%%
\label{sec:tag}
The tag net consists of two parts Context Extractor \& Tag Embedding Net. This is illustrated in figure~\ref{fig:tag}.

\textbf{Extract Context}:
The first step is to extract the caption of the image using NeuralTalk2 \cite{Karpathy_NIPS2014} model. We find the part-of-speech(POS) tag present in the caption. POS taggers have been developed for two well known corpuses, the Brown Corpus and the Penn Treebanks. For our work, we are using the Brown Corpus tags. The tags are clustered into three category namely Noun tag, Verb tag and Question tags (What, Where, \dots). Noun tag consists of all the noun \& pronouns present in the caption sentence and similarly, verb tag consists of verb \& adverbs present in the caption sentence. The question tags consists of the 7-well know question words {i.e., why, how, what, when, where, who and which}. Each tag token is represented as a one-hot vector of the dimension of vocabulary size. For generalization, we have considered 5 tokens from each category of the Tags.

\textbf{Tag Embedding Net}:
The embedding network consists of word embedding followed by temporal convolutions neural network followed by max-pooling network. In the first step, sparse high dimensional one-hot vector is transformed to dense low dimension vector using word embedding. After this, we apply temporal convolution on the word embedding vector. The uni-gram, bi-gram and tri-gram feature are computed by applying convolution filter of size 1, 2 and 3 respectability. Finally, we applied max-pooling on this to get a vector representation of the tags as shown figure~\ref{fig:tag}. We concatenated all the tag words followed by fully connected layer to get feature dimension of 512. We also explored joint networks based on concatenation of all the tags, on element-wise addition and element-wise multiplication of the tag vectors. However, we observed that convolution over max pooling and joint concatenation gives better performance based on CIDer score.
\[F_C =\text{Tag\_CNN}(C_t)\]
Where, T\_CNN is Temporally Convolution Neural Network applied on word embedding vector with kernel size three.

%%%%%%%%%%%%%%%%%%%%%%%%%%%%%%%%%%%%%%%%%%%%%%%%%%%%%%%%%%%%%%%%%%%%%%%
\subsection{Place net}
%%%%%%%%%%%%%%%%%%%%%%%%%%%%%%%%%%%%%%%%%%%%%%%%%%%%%%%%%%%%%%%%%%%%%%%
Visual object and scene recognition plays a crucial role in the image. Here, places in the image are  labeled with scene semantic categories\cite{Zhou_PAMI2017}, comprise of large and diverse type of environment in the world, such as (amusement park, tower, swimming pool, shoe shop, cafeteria, rain-forest, conference center, fish pond, etc.). So we have used different type of scene semantic categories present in the image as a place based context to generate natural question. A place365 is a convolution neural network is modeled to classify 365 types of scene categories, which is trained on the place2 dataset consist of 1.8 million of scene images. We have used a pre-trained VGG16-places365 network to obtain place based context embedding feature for  various type scene categories present in the image. The context feature $F_C$ is obtained by:
 \[F_C =w*\text{p\_conv}(I) +b \]
Where, $\text{p\_conv}$ is Place365\_CNN. We have extracted $conv5$ features of dimension 14x14x512 for attention model and FC8 features of dimension 365 for joint, addition and hadamard model of places365. Finally, we use a linear transformation to obtain a 512 dimensional vector.
 
We explored using the CONV$5$ having feature dimension 14x14 512, FC$7$ having 4096 and FC8 having feature dimension of 365 of places365.

% \subsection{Analysis of Network parameters}
% \subsection{Analysis of Context}

\section{Ablation Analysis}\label{disc}
\subsection{Sampling Exemplar: KNN vs ITML}
Our method is aimed at using efficient exemplar-based retrieval techniques. 
We have experimented with various exemplar methods, such as  ITML \cite{davis_ACM2007} based metric learning for image features and KNN based approaches. We observed KNN based approach (K-D tree) with Euclidean metric is a efficient method for finding exemplars. Also we observed that ITML is computationally expensive and also depends on the training procedure. The table provides the experimental result for Differential Image Network variant with k (number of exemplars) = 2  and Hadamard method:	

 \begin{table}[h!]
\scriptsize
\centering
\begin{tabular}{|l|c|cccc|}
\hline \bf Meth & \bf Exemplar & \bf BLEU-1 & \bf Meteor & \bf Rouge  & \bf CIDer \\ \hline
KNN &IE(K=2)& 23.2 &8.9 & \textbf{27.8} & 22.1\\ 
ITML &IE(K=2)& 22.7 &9.3 & 24.5 & 22.1\\ \hline
\end{tabular}
\caption{\label{score_tab_11a}VQG-COCO-dataset, Analysis of different methods of finding Exemplars for hadamard model. ITML vs KNN based methods. We see that both give more or less similar results but since ITML is computationally expensive and the dataset size is also small, it is not that efficient for our use. All these experiment are for the differential image network for K=2 only.}
\end{table}

\subsection{Question Generation approaches: Sampling vs Argmax}
 We obtained the decoding using standard practice followed in the literature~\cite{Sutskever_NIPS2014}. This method selects the argmax sentence. Also, we evaluated our method by sampling from the probability distributions and provide the results for our proposed MDN-Joint method for VQG dataset as follows:
 \begin{table}[h!]
\scriptsize
\centering
\begin{tabular}{|l|cccc|}
\hline \bf Meth &\bf BLEU-1 & \bf Meteor & \bf Rouge  & \bf CIDer \\ \hline
Sampling &  17.9 &11.5 & 20.6 & 22.1\\ 
Argmax &  36.0 &23.4 & 41.8 & 50.7\\ \hline
\end{tabular}
\caption{\label{score_tab_12}VQG-COCO-dataset, Analysis of question generation approaches:sampling vs Argmax in MDN-Joint model for K=5 only. We see that Argmax clearly outperforms the sampling method.}
\end{table}

\subsection{How are exemplars improving Embedding}
In Multimodel differential network, we use exemplars and train them using a triplet loss. It is known that using a triplet network, we can learn a representation that accentuates how the image is closer to a supporting exemplar as against the opposing exemplar~\cite{Hoffer_Springer2015,Frome_ICCV2007}. The Joint embedding is obtained between the image and language representations. Therefore the improved representation helps in obtaining an improved context vector. Further we show that this also results in improving VQG. %More analysis in terms of understanding how the methods qualitatively improves attention is included  in the supplementary material.

\subsection{Are exemplars required?}
We had similar concerns and validated this point by using random exemplars for the nearest neighbor for MDN. (k=R in table~\ref{score_tab_11}) In this case the method is similar to the baseline. This suggests that with random exemplar, the model learns to ignore the cue. 
%\subsection{Why does even using random exemplar improve over the baseline?}
%The improvement over the baseline is slight (0.3\%). One reason is that we do have a triplet model that is learnt over the random exemplars to obtain attention that is trained to minimize the VQG objective. Therefore, the learned model may be contributing an additional marginal improvement. However, as can be seen from the results, using a systematic exemplar approach is much more influential (~4\% improvement) over the baseline.
% the results are present in supplementary material.

\subsection{Are captions necessary for our method?}
 This is not actually necessary. In our method, we have used an existing image captioning method~\cite{Karpathy_CVPR2015} to generate captions for images that did not have them. For VQG dataset, captions were available and we have used that, but, for VQA dataset captions were not available and we have generated captions while training. We provide detailed evidence with respect to caption-question pairs to ensure that we are generating novel questions. While the caption generates scene description, our proposed method generates semantically meaningful and novel questions.
 Examples for Figure 1 of main paper:
First Image:- Caption- A young man skateboarding around little cones. Our Question- Is this a skateboard competition?
Second Image:- Caption- A small child is standing on a pair of skis.
Our Question:- How old is that little girl?
% 6th Image of Figure 5: Caption- The room is filled with lots of teddy bears. Our Question:- How long have you collected bears?

\subsection{Intuition behind Triplet Network:}
The intuition behind use of triplet networks is clear through this paper\cite{Frome_ICCV2007} that first advocated its use. The main idea is that when we learn distance functions that are “close” for similar and “far” from dissimilar representations, it is not clear that close and far are with respect to what measure. By incorporating a triplet we learn distance functions that learn that  “A is more similar to B as compared to C”. Learning such measures allows us to bring target image-caption joint embeddings that are closer to supporting exemplars as compared to contrasting exemplars.

% \section{Discussion}
% In  this  section  we  further  discuss  different  aspects  of our  method  that  are  useful  for  understanding  the  method in more detail

% \subsection{How likely is it that you will find the exemplars that satisfy their properties?}
%  We had similar concerns and validated this point in the paper by using random exemplars for the nearest neighbor for MDN. In this case the method is similar to the baseline. This suggests that with random exemplar, the model learns to ignore the cue.
% \subsection{Is multimodal embedding dominated by the caption or by the image or both?}
% We had considered this issue and had conducted experiments by considering image only, by considering caption only, and by considering nearest neighbor using the semantic feature of both image and caption. We observed that the joint image-caption embedding from the baseline model performed better than other two. Therefore we believe that both contribute to the embedding. 
\section{Analysis of Network}
\subsection{Analysis of Tag Context}
\label{sec:context_analysis_tag}

Tag is language based context. These tags are extracted from caption, except question-tags which is fixed as the 7 'Wh words' (What, Why, Where, Who, When, Which and How). We have experimented with Noun tag, Verb tag and 'Wh-word' tag as shown in tables. Also, we have experimented in each tag by varying the number of tags from 1 to 7. We combined different tags using 1D-convolution, concatenation, and addition of all the tags and observed that the concatenation mechanism  gives better results.  

As we can see in the table~\ref{score_tab_7} that taking Nouns, Verbs and Wh-Words as context, we achieve significant improvement in the BLEU, METEOR and CIDEr scores from the basic models which only takes the image and the caption respectively.
Taking Nouns generated from the captions and questions of the corresponding training example as context, we achieve an increase of 1.6\% in Bleu Score and 2\% in METEOR and 34.4\% in CIDEr Score from the basic Image model. Similarly taking Verbs as context gives us an increase of 1.3\% in Bleu Score and 2.1\% in METEOR and 33.5\% in CIDEr Score from the basic Image model. And the best result comes when we take 3 Wh-Words as context and apply the Hadamard Model with concatenating the 3 WH-words. \\
Also in Table ~\ref{score_tab_9} we have shown the results when we take more than one words as context. Here we show that for 3 words i.e 3 nouns, 3 verbs and 3 Wh-words, the Concatenation model performs the best. In this table the conv model is using 1D convolution to combine the tags and the joint model combine all the tags.
\label{sec:context_analysis}
% %-------------------------------------------------------------------------

\subsection{Analysis of Context: Exemplars }
\label{sec:model_analysis}
In Multimodel Differential Network and Differential Image Network, we use exemplar images(target, supporting and opposing image) to obtain the differential context. We have performed the experiment based on the single exemplar(K=1), which is one supporting and one opposing image along with target image, based on two exemplar(K=2), i.e. two supporting and two opposing image along with single target image. similarly we have performed experiment for K=3 and K=4 as shown in table-~\ref{score_tab_11}.

 \subsection{Mixture Module: Other Variations} Hadamard method uses element-wise multiplication whereas {Addition method} uses element-wise addition in place of the concatenation operator of the Joint method. The Hadamard method finds a correlation between image feature and caption feature vector while the Addition method learns a resultant vector.
In the attention method, the output $S_{i}$ is the weighted average of attention probability vector $P_{att}$ and convolutional features $G_{img}$. The attention probability vector weights the contribution of each convolutional feature based on the caption vector. This attention method is similar to work stack attention method~\cite{Yang_CVPR2016}. The attention mechanism is given by:
\begin{equation}
    \begin{split}
        & h_{att}= \tanh({W_I}{G_{img}} \oplus ({W_C}{F_{cap}}+{b_c})) \\
        & P_{att}= \mbox{Softmax}({W^T_P}{h_{att}}+{b_P}) \\
        & V_{att}= \sum_{i}{P_{att}(i)}{G_{img}(i)}\\
        & A_{att}= V_{att} + f_{i} \\
        & s_{i}=\tanh({W_A} A_{att} + b_A)
    \end{split}
\end{equation}
where $G_{img}$ is  the 14x14x512-dimensional  convolution feature map from the fifth convolution layer of VGG-19 Net of image $X_{i}$ and $f_{i}$ is the caption context vector. The attention probability vector $P_{att}$ is a 196-dimensional vector. $W_{I},W_{C},W_{P}$ are the weights and $b_c, b_A, b_c $ is the bias for different layers. We evaluate the different approaches and provide results for the same. Here $\oplus$ represents element-wise addition.

\subsection{Evaluation Metrics}
Our task is similar to encoder -decoder framework of machine translation. we have used same evaluation metric is used in machine translation. BLEU\cite{Papineni_ACL2002} is the first metric to find the correlation between generated question with ground truth question. BLEU score is used to measure the precision value, i.e That is how much words in the predicted question is appeared in reference question. BLEU-n score measures the n-gram precision for counting co-occurrence on reference sentences. we have evaluated BLEU score from n is 1 to 4. The mechanism of ROUGE-n\cite{Lin_ACL2004} score  is similar to BLEU-n,where as, it measures recall value instead of precision value in BLEU. That is how much words in the reference question is appeared in predicted question.Another version ROUGE metric is ROUGE-L, which  measures longest common sub-sequence present in the generated question. METEOR\cite{Banerjee_ACL2005} score is another useful evaluation metric to calculate the similarity between generated question with reference one by considering synonyms, stemming and paraphrases. the output of the METEOR score measure the word matches between predicted question and reference question. In VQG, it compute the word match score between predicted question with five reference question. CIDer\cite{Vedantam_CVPR2015} score is a consensus based evaluation metric.  It measure human-likeness, that is the sentence is written by human or not. The consensus is measured, how often n-grams in the predicted question are appeared in the reference question. If the n-grams in the predicted question sentence is appeared more frequently in reference question then question is less informative and have low CIDer score. We provide our results using all these metrics and compare it with existing baselines.

% \bibliography{emnlp2018}
% \bibliographystyle{acl_natbib_nourl}

% \end{document}

\end{document}